\documentclass[journal,final]{IEEEtran}

\usepackage{cite}
\usepackage{graphicx}
\usepackage{amsmath}
\usepackage{amssymb}
\usepackage{booktabs}
\usepackage[caption=false,font=footnotesize]{subfig}
\usepackage{hyperref}

\usepackage{xcolor,comment}
\hypersetup{
    colorlinks,
    linkcolor={black},
    citecolor={black},
    urlcolor={cyan}
}

\begin{document}
\title{Forensic Similarity for Digital Images}

\author{Owen~Mayer,~\IEEEmembership{Student Member,~IEEE,}
        Matthew~C.~Stamm,~\IEEEmembership{Member,~IEEE}
\thanks{O. Mayer and M.C. Stamm are with the Department
of Electrical and Computer Engineering, Drexel University, Philadelphia,
PA, 19104 USA \mbox{e-mail}: om82@drexel.edu, mstamm@drexel.edu.}
\thanks{This material is based upon work supported by the National Science
Foundation under Grant No. 1553610.}}

\maketitle

\newcommand\blfootnote[1]{%
  \begingroup
  \renewcommand\thefootnote{}\footnote{#1}%
  \addtocounter{footnote}{-1}%
  \endgroup
}

\begin{abstract}
In this paper we introduce a new digital image forensics approach called forensic similarity, which determines whether two image patches contain the same forensic trace or different forensic traces. One benefit of this approach is that prior knowledge, e.g. training samples, of a forensic trace are not required to make a forensic similarity decision on it in the future. 
To do this, we propose a two part deep-learning system composed of a CNN-based feature extractor and a three-layer neural network, called the similarity network. This system maps pairs of image patches to a score indicating whether they contain the same or different forensic traces. We evaluated system accuracy of determining whether two image patches were 1)~captured by the same or different camera model, 2)~manipulated by the same or different editing operation, and 3)~manipulated by the same or different manipulation parameter, given a particular editing operation. Experiments demonstrate applicability to a variety of forensic traces, and importantly show efficacy on ``unknown" forensic traces that were not used to train the system. Experiments also show that the proposed system significantly improves upon prior art, reducing error rates by more than half. Furthermore, we demonstrated the utility of the forensic similarity approach in two practical applications: forgery detection and localization, and database consistency verification.
\end{abstract}


\section{Introduction}
\IEEEPARstart{T}{rustworthy} multimedia content is important to a number of institutions in today's society, including news outlets, courts of law, police investigations, intelligence agencies, and social media websites. As a result, multimedia forensics approaches have been developed to expose tampered images, determine important information about the processing history of images, and identify details about the camera make, model, and device that captured them. These forensic approaches operate by detecting the visually imperceptible traces, or ``fingerprints,'' that are intrinsically introduced by a particular processing operation~\cite{stamm2013overview}. 

Recently, researchers have developed deep learning methods that target digital image forensic tasks with high accuracy. For example, convolutional neural network (CNN) based systems have been proposed that accurately detect traces of median filtering~\cite{chen2015median}, resizing~\cite{bayar2017robustness,bunk2017detection}, inpainting~\cite{zhu2018deep}, multiple processing operations~\cite{bayar2016deep,bayar2018mislnet,boroumand2018deep}, processing order~\cite{bayar2018order}, and double JPEG compression~\cite{barni2017aligned,amerini2017localization}. Additionally, researchers have proposed approaches to identify the source camera model of digital images~\cite{tuama2016camera,bondi2017preliminary,bondi2017first,bayar2017augmented}, and identify their origin social media website~\cite{amerini2017tracing}.\looseness=-1

However, there are two main drawbacks to many of these existing approaches. First is that many deep learning systems assume a closed set of forensic traces, i.e. known and closed set of possible editing operations or camera models. That is, these methods require prior training examples from a particular forensic trace, such as the source camera model or editing operation, in order to identify it again in the future. This requirement is a significant problem for forensic analysts, since they are often presented with new or previously unseen forensic traces. 
Additionally, it is often not feasible to scale deep learning systems to contain the large numbers of classes that a forensic investigator may encounter. For example, systems that identify the source camera model of an image often require hundreds of scene-diverse images per camera model for training~\cite{bondi2017preliminary}. To scale such a system to contain hundreds or thousands of camera models requires a prohibitively large data collection undertaking.\looseness=-1 

A second drawback of these existing approaches is that many forensic investigations do not require explicit identification of a particular forensic trace. For example, when analyzing a splicing forgery, which is a composite of content from multiple images, it is often sufficient to simply detect a region of an image that was captured by a different source camera model, without explicitly identifying that source. 
That is, the investigator does not need to determine the exact processing applied to the image, or the true sources of the pasted content, just that an inconsistency exists within the image. 
In another example, when verifying the consistency of an image database, the investigator does not need to explicitly identify which camera models were used to capture the database, just that only one camera model or many camera models were used.\blfootnote{Code and pre-trained CNNs for this project are available at \href{http://www.gitlab.com/MISLgit/forensic-similarity-for-digital-images}{gitlab.com/MISLgit/forensic-similarity-for-digital-images} and our laboratory website \href{http://misl.ece.drexel.edu/downloads/}{misl.ece.drexel.edu/downloads/}.}




Recently, researchers have proposed CNN-based forensic systems for digital images that do not require a closed and known set of forensic traces. Research in~\cite{bayar2018open} proposed a system 
to output a binary decision indicating whether an image was captured by a camera model used for training, or from an unknown camera model that was not used in training. 
Research in~\cite{bondi2017tampering} showed that features learned by a CNN for source camera model identification can be iteratively clustered to identify spliced images. In addition, other CNN-based methods have been proposed for forgery localization~\cite{cozzolino2018camera,cozzolino2018noiseprint,huh2018forensics}. The authors showed that these type systems can detect spliced images even when the camera models were not used to train the system. 

In this paper, we propose a new digital image forensics approach that operates on an open set of forensic traces. This approach, which we call \textit{forensic similarity}, determines whether two image patches contain the \textit{same or different} forensic traces. This approach is different from other forensics approaches in that it does not explicitly identify the particular forensic traces contained in an image patch, just whether they are consistent across two image patches. The benefit of this approach is that prior knowledge of a particular forensic trace is not required to make a similarity decision on it.

To do this, we propose a two part deep learning system. In the first part, called the feature extractor, we use a CNN to extract general low-dimensional forensic features, called deep features, from an image patch. Prior research has shown that CNNs can be trained to map image patches onto a low-dimensional feature space that encodes general and high-level forensic information about that patch~\cite{bayar2018open,bondi2017tampering,bondi2017first, mayer2018similarity,bayar2017robustness,mayer2018unified}. Next, we use a three layer neural network to map pairs of these deep features onto a similarity score, which indicates whether the two image patches  contain the same forensic trace or different forensic traces.

We experimentally evaluate the efficacy of our proposed approach in several scenarios. We evaluate the performance of our proposed forensic similarity system at determining whether two image patches were 1) captured by the same or different camera model, 2) manipulated the same or different editing operation, and 3) manipulated same or different manipulation parameter, given a particular editing operation. Importantly, we evaluate performance on camera models, manipulations and manipulation parameters not used during training, demonstrating that this approach is effective in open-set scenarios.

Furthermore, we demonstrate the utility of this approach in two practical applications that a forensic analyst may encounter. In the first application, we demonstrate that our forensic similarity system detects and localizes image forgeries. Since image forgeries are often a composite of content captured by two camera models, these forgeries are exposed by detecting the image regions that are forensically dissimilar to the host image. In the second application, we show that the forensic similarity system verifies whether a database of images was captured by all the same camera model, or by different camera models. This is useful for flagging social media accounts that violate copyright protections by stealing content, often from many different sources.

This paper is an extension of our earlier work in~\cite{mayer2018similarity}. In our previous work, we proposed a proof-of-concept similarity system for comparing the source camera model of two image patches and evaluated on a limited set of camera models. This system consisted of a pair of CNNs in a ``Siamese'' configuration, and a shallow neural network.
In this work, we extend our previous work~\cite{mayer2018similarity} in several ways. First, we reframe the approach as a general system that is applicable to any measurable forensic trace, such as manipulation type or editing parameter, not just source camera model. Second, we significantly improve the system architecture and training procedure. We show these improvements lead to an over 50\% reduction classification error for camera model comparisons relative to the method described in~\cite{mayer2018similarity}. Among several changes, significant improvements are due to the utilization of full RGB color images, instead of relying soley on the green color channel, as well as allowing the entire network to update in the secondary learning phase. Finally, we experimentally evaluate our proposed approach in an vastly expanded range of scenarios, and demonstrate utility in two practical applications.

In addition to our prior work in~\cite{mayer2018similarity}, other researchers have utilized Siamese network configurations for multimedia forensics. Research in~\cite{cozzolino2018noiseprint} and~\cite{cozzolino2018camera} proposes an image transformation system that extracts a noise residual map, called the ``Noiseprint,'' induced by camera model processes. During training, pairs of extracted residual maps from the same camera model and image position are encouraged to have small pixel-by-pixel Euclidean distances. Unlike our proposed system, their approach inputs a single image and outputs a noise residual map, and then searches this map for forgery related inconsistencies. 

In other research~\cite{huh2018forensics}, a CNN in Siamese configuration is trained with pairs of image patches and labels associated with EXIF header information of an image. Their approach produces a ``self-consistency'' heatmap, which highlights image regions that have undergone forgery~\cite{huh2018forensics}. They show state-of-the-art forgery detection and localization results. In Sec.~\ref{sec:splicing_experiments}, we compare against their approach for forgery detection.

The remaining parts of the manuscript are outlined as follows. In Sec.~\ref{sec:formulation}, we motivate and formalize the concept of forensic similarity. In Sec.~\ref{sec:approach}, we detail our proposed deep-learning system implementation and training procedure. In this section, we describe how to build and train the CNN-based feature extractor and similarity network. In Sec.~\ref{sec:experiments1}, we evaluate the effectiveness of our proposed approach in a number of forensic situations, and importantly effectiveness on unknown forensic traces. Finally, in Sec.~\ref{sec:experiments2}, we demonstrate the utility of this approach in two practical applications. 
\begin{figure}
\centering
\includegraphics[width=0.91\linewidth]{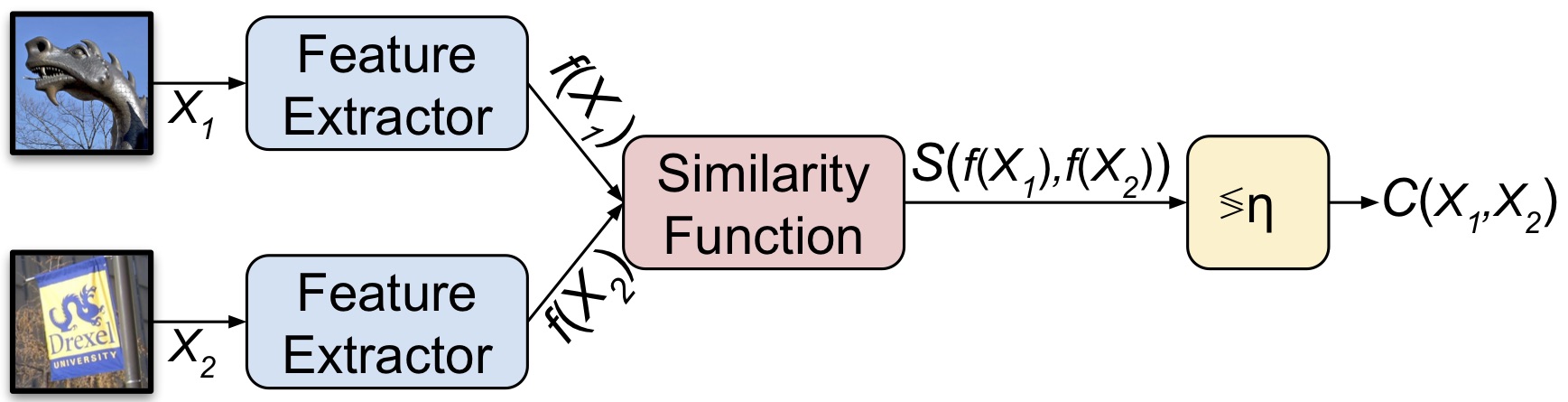}\vspace{-1mm}
\caption{Forensic similarity system overview.}
\label{fig:sys_overview}
\vspace{-4.5mm}
\end{figure}

\definecolor{myOrange}{rgb}{1.0, 0.4980392156862745, 0.054901960784313725} 
\definecolor{myBlue}{rgb}{0.12156862745098039, 0.4666666666666667, 0.7058823529411765,}

\begin{figure*}[t]
\captionsetup[subfigure]{justification=centering}
\null\hfill
\subfloat[Reference Image \newline (Google Pixel 1)]{\includegraphics[height=1.6in]{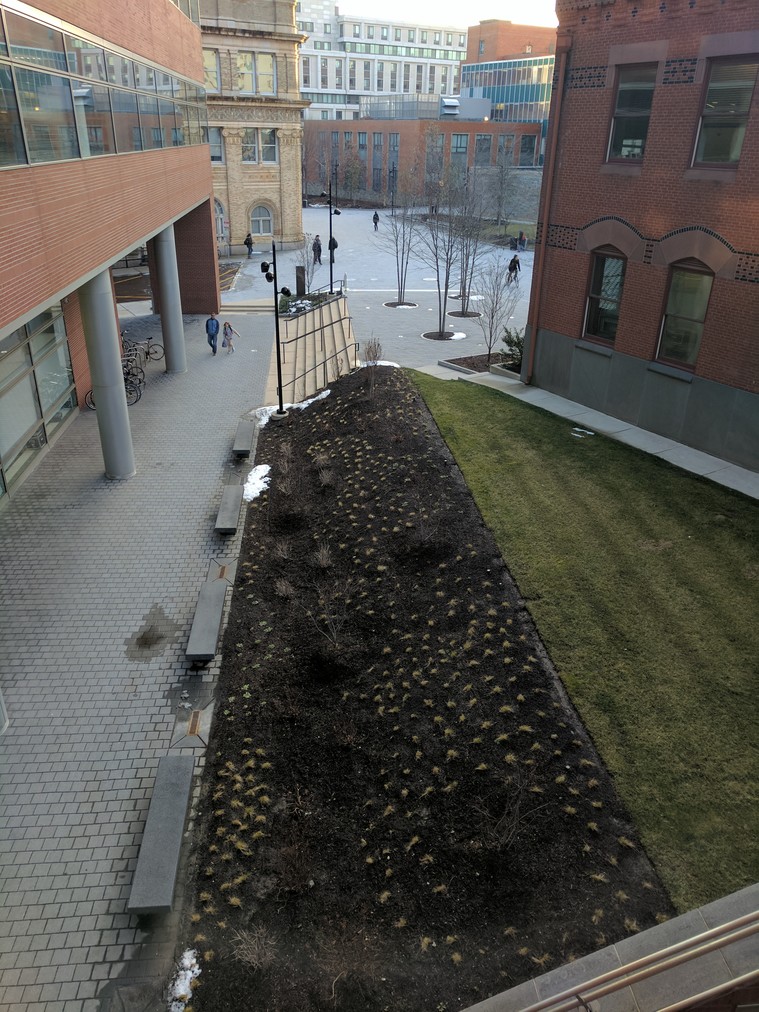}}
\hfill
\subfloat[Google Pixel 1 Image]{\fcolorbox{white}{myBlue}{\includegraphics[height=1.6in]{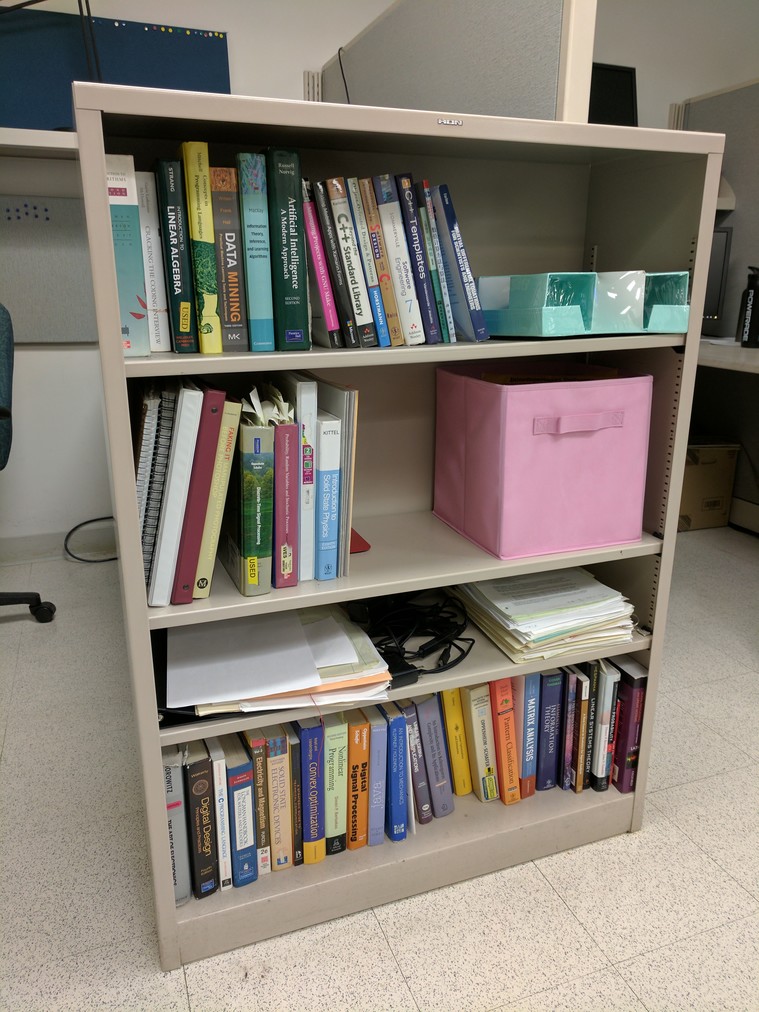}}}
\hfill
\subfloat[Asus Zenfone 3 Image]{\fcolorbox{white}{myOrange}{\includegraphics[height=1.6in]{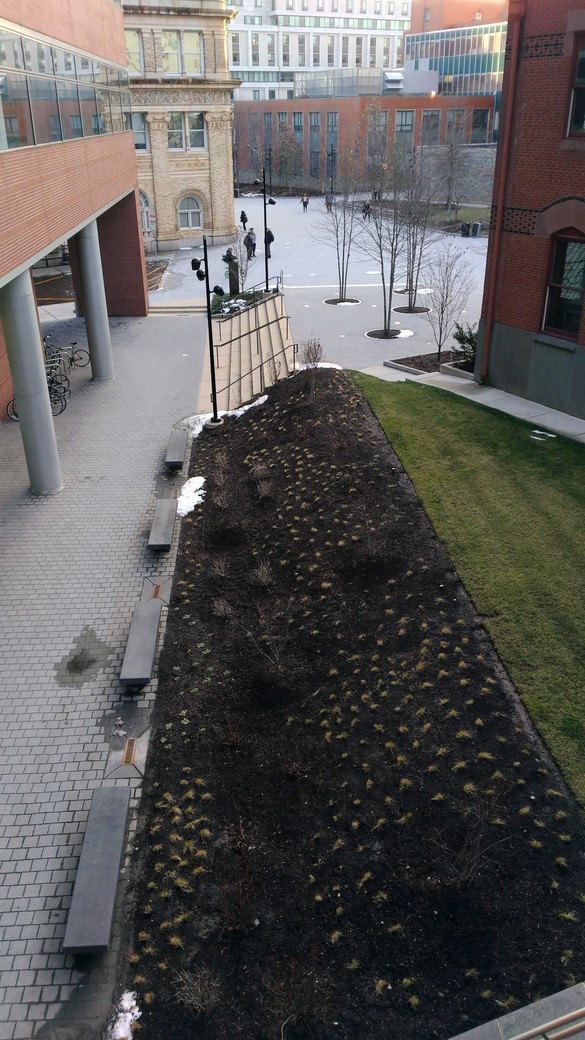}}}
\hfill
\subfloat[Forensic Similarity Distribution]{\includegraphics[height=1.6in]{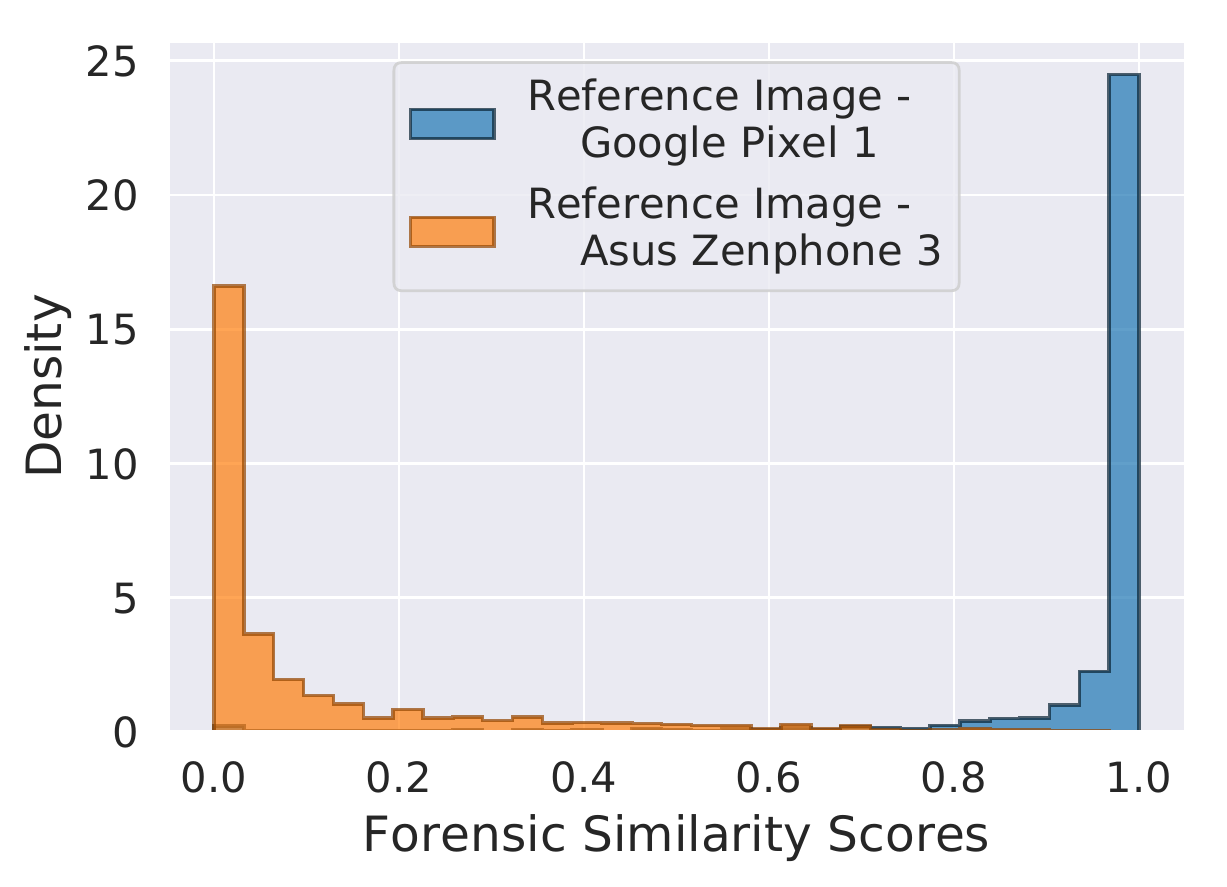}}
\hfill
\hfill\null
\caption{Camera-model based forensic similarity scores measured between random pairings of patches from the reference image (a), captured by a Google Pixel 1 smartphone rear-facing camera, and from (b) a different Google Pixel 1 image or (c) a Asus Zenfone 3 image of the same scene as the reference image. The histogram (d) shows the distribution of these forensic similarity scores. Forensic similarity between patches from the two Google Pixel images is high, even though the two images depict different scenes and content, and were captured in different lighting environments. The forensic similarity between patches from the Google Pixel and Asus Zenfone cameras is low, even though the two images depict similar content. }
\label{fig:sim_example}
\vspace{-3mm}
\end{figure*}

\section{Forensic Similarity}
\label{sec:formulation}
Prior multimedia forensics approaches for digital images have focused on identifying or classifying a particular forensic trace (e.g. source camera model, processing history) in an image or image patch. These approaches, however, suffer from two major drawbacks in that 1) training samples from a particular trace are required to identify it, and 2) not all forensic analyses require identification of a trace. For example, to expose a splicing forgery it is sufficient to identify that the forged image is simply composite content from different sources, without needing to explicitly identify those sources.

In this paper, we propose a new general approach that addresses these drawbacks. We call this approach \textit{forensic similarity}. Forensic similarity is an approach that determines if two image patches have the same or different forensic trace. Unlike prior forensic approaches, it does not identify a particular trace, but still provides important forensic information to an investigator. The main benefit of this type of approach is that it is able to be practically implemented in open-set scenarios. That is, a forensic similarity based system does not inherently require training samples from a forensic trace in order to make a forensic similarity decision. Later, in Sec.~\ref{sec:approach}, we describe how this approach is implemented using a CNN-based deep learning system.\looseness=-1

In this work, we define a \textit{forensic trace} to be a signal embedded in an image that is induced by, and captures information about, a particular signal processing operation performed on the image. Forensic traces are inherently unrelated to the perceptual content of the image; two images depicting different scenes may contain similar forensic traces, and two images depicting similar scenes may contain different forensic traces.
Common mechanisms that induce a forensic trace in an image are: the camera model that captured the image, the social media website where the image was downloaded from, and the processing history of the image. A number of approaches have been researched to extract and identify the forensic traces related to these mechanisms~\cite{bayar2018mislnet,bondi2017first,bayar2017generic}. 

These prior approaches, however, assume a closed set of forensic traces.
They are designed to perform a mapping \mbox{$\mathbb{X} \to \mathbb{Y}$} where $\mathbb{X}$ is the space of image patches and  $\mathbb{Y}$ is the space of known forensic traces that are used to train the system, e.g. camera models or editing operations. However when an input image patch has a forensic trace $y \notin \mathbb{Y}$, the identification system is still forced to map to the space $\mathbb{Y}$, leading to an erroneous result. That is, the system will misclassify this new  ``unknown" trace as a ``known" one in $\mathbb{Y}$.

This is problematic since in practice forensic investigators are often presented with images or image patches that contain a forensic trace for which the investigator does not have training examples. We call these \textit{unknown forensic traces.} This may be a camera model that does not exist in the investigator's database, or previously unknown editing operation. In these scenarios, it is still important to glean important forensic about the image or image patch.

To address this, we propose a system that is capable of operating on unknown forensic traces. Instead of building a system to identify a particular forensic trace, we ask the question ``do these two image patches contain the same forensic trace?" Even though the forensic similarity system may have never seen a particular forensic trace before, it is still able to distinguish whether they are the same or different across two patches. This type of question is analogous to the content-based image retrieval problem~\cite{wan2014deep}, and the speaker verification problem~\cite{heigold2016end}. 

We define forensic similarity as the function
\begin{equation}
C: \mathbb{X}\times\mathbb{X} \to \left\lbrace 0, 1 \right\rbrace,
\end{equation}
that compares two image patches. This is done by mapping two input image patches $X_1, X_2 \in \mathbb{X}$ to a score indicating whether the two image patches have the same or different forensic trace. A score of 0 indicates the two image patches contain different forensic traces, and a score of 1 indicates they contain the same forensic trace. In other words
\begin{align}
C(X_1,X_2) = \begin{cases}
0 & \text{if $X_1,X_2$ diff. forensic traces,}\\
1 & \text{if $X_1,X_2$ same forensic trace.}
\end{cases}
\end{align}
To construct this system, we propose a forensic similarity system consisting of two main conceptual parts, which are shown in the system overview in Fig.~\ref{fig:sys_overview}. The first conceptual part is called the feature extractor
\begin{equation}
f: \mathbb{X} \to \mathbb{R}^{N},
\label{eq:feature_extractor}
\end{equation}
which maps an input image patch $X$ to a real valued N-dimensional feature space. This feature space encodes high-level forensic information about the image patch $X$. Recent research in multimedia forensics has shown that convolutional neural networks (CNNs) are powerful tools for extracting general, high-level forensic information from image patches~\cite{mayer2018unified}. We specify how this is done in Sec.~\ref{sec:approach}, where we describe our proposed implementation of the forensic similarity system.

Next we define the second conceptual part, the similarity function 
\vspace{-2mm}
\begin{equation}
S: \mathbb{R}^{N}\times\mathbb{R}^{N} \to \left[0,1\right],
\label{eq:similarity_map}\vspace{-1mm}
\end{equation}
that maps pairs of forensic feature vectors to a \textit{similarity score} that takes values from 0 to 1. A low similarity score indicates that the two image patches $X_1$ and $X_2$ have dissimilar forensic traces, and a high similarity score indicates that the two forensic traces are highly similar.

Finally, we compare the similarity score $S(f(X_1),f(X_2))$ of two image patches $X_1$ and $X_2$ to a threshold $\eta$ such that
\begin{equation}
C(X_1,X_2) = \begin{cases}
0 & \text{if } S(f(X_1),f(X_2)) \leq \eta \\
1 & \text{if } S(f(X_1),f(X_2)) > \eta
\end{cases}.
\label{eq:overallsystem}
\end{equation}
In other words, the proposed forensic similarity system takes two image patches $X_1$ and $X_2$ as input. A feature extractor maps these two input image patches to a pair of feature vectors $f(X_1)$ and $f(X_2)$, which encode high-level forensic information about the image patches. 
Then, a similarity function maps these two feature vectors to a similarity score, which is then compared to a threshold. 
A similarity score above the threshold indicates that $X_1$ and $X_2$ have the same forensic trace (e.g. processing history or source camera model), and a similarity score below the threshold indicates that they have different forensic traces.

Examples of forensic similarity scores are shown in Fig.~\ref{fig:sim_example}. In this example, we calculated forensic similarity scores between small patches randomly chosen from three different images: two captured by a Google Pixel 1, and one captured by an Asus Zenfone 3. Neither of these camera models were used to train the system. When both patches were captured by the same camera model, forensic similarity scores are high and near one, as shown in by the blue distribution in Fig.~\ref{fig:sim_example}(d). When both patches were captured by different camera models, the forensic similarity scores are low and near zero, as shown by the orange curve. An important quality to note is that forensic similarity is invariant to the semantic content depicted in the image. For example, even though image (a) and image (c) depict very similar scenes their forensic similarity is low since they were captured by different camera models. This is an important distinction from computer vision approaches such as object detection and scene recognition, which are invariant to any non-content related qualities such as source camera model.\looseness=-1

\begin{figure*}[t]
\centering
\includegraphics[width=0.975\linewidth]{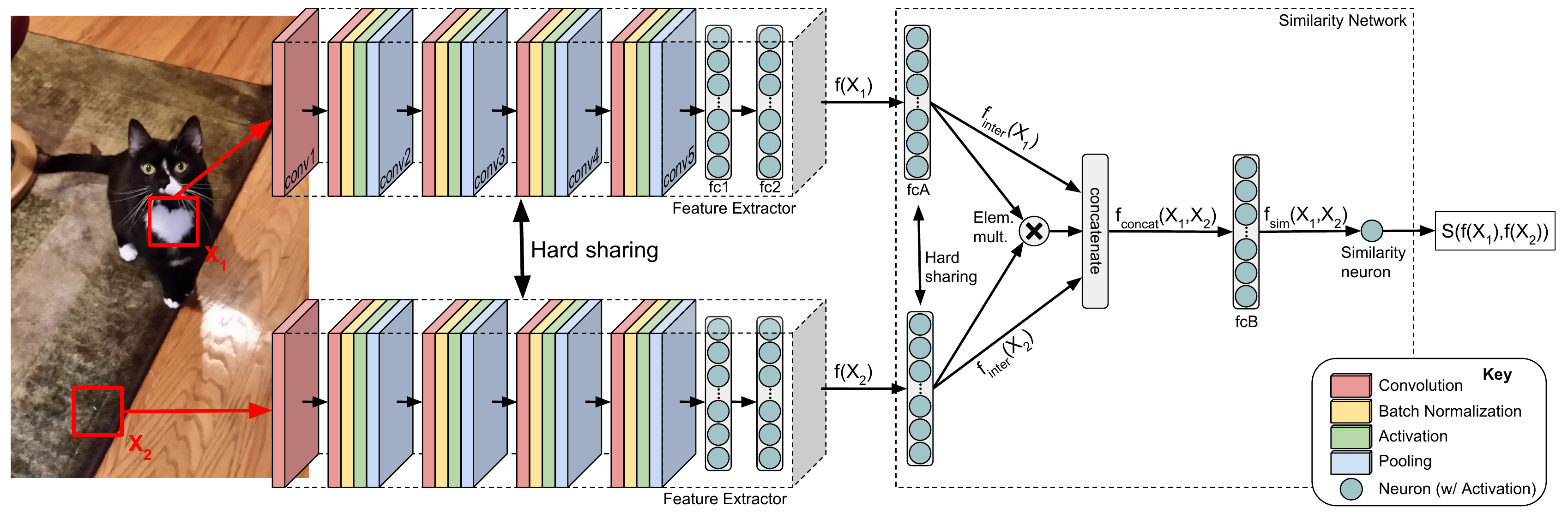}\vspace{-4mm}
\caption{The neural network architecture of the proposed forensic similarity system. The system is composed of a pair of CNN-based feature extractors, in a hard sharing (Siamese) configuration, which feed low-dimensional, high-level forensic feature vectors to the similarity network. The similarity network is a neural network that maps feature vectors from two image patches to a similarity score indicating whether they contain the same or different forensic traces.}
\label{fig:full_arch}
\vspace{-4mm}
\end{figure*}

\section{Proposed Approach}
\label{sec:approach}

In this section, we describe our proposed deep learning system architecture and associated training procedure for forensic similarity. 
In our proposed forensic similarity architecture and training procedure, we build upon prior CNN-based techniques used in multimedia forensics literature, as well as propose a number of innovations that are aimed at extracting robust forensic features from images and accurately determining forensic similarity between two image patches.

Our proposed forensic similarity system consists of two conceptual elements: 1) a CNN-based feature extractor that maps an input image onto a low-dimensional feature space that encodes high level forensic information, and 2) a three-layer neural network, which we call the similarity network, that maps pairs of these features to a score indicating whether two image patches contain the same forensic trace. This system is trained in two successive phases. In the first phase, called \textit{Learning Phase A}, we train the feature extractor. In the second phase, called \textit{Learning Phase B} we train the similarity network. Finally, in this section we describe an entropy-based method of patch selection, which we use to filter out patches that are not suitable for forensic analysis.

\vspace{-2mm}
\subsection{Learning Phase A - Feature Extractor}
Here we describe the deep-learning architecture and training procedure of the feature extractor that maps an input image patch onto a low dimensional feature space, which encodes forensic information about the patch. This is the mapping described by~\eqref{eq:feature_extractor}. Later, pairs these feature vectors are used as input to the similarity network described below in Sec.~\ref{sec:approach:ssec:similarity}.

Developments in machine learning research have shown that CNNs are powerful tools when used as generic feature extractors. This is done by robustly training a deep convolutional neural network for a particular task, and then using the neuron activations, at a deep layer in the network, as a feature representation of an image~\cite{donahue2014decaf}. These neuron activations are called ``deep features," and are often extracted from the last fully connected layer of a network. Research has shown that deep features extracted from a CNN trained for object recognition tasks can be used to perform scene recognition tasks~\cite{zhou2014learning} and remote sensing tasks~\cite{penatti2015deep}.

In multimedia forensics research, it has been shown that deep features based approaches also very powerful for digital image forensics tasks. For example, work in~\cite{bondi2017first} showed that deep features from a network trained on one set of camera models can be used to train an support vector machine to identify a different set of camera models. Work in~\cite{bayar2018open} showed that deep features from a CNN can be used to determine whether an image was captured by a camera model used during training. Furthermore, it has been shown that deep features from a CNN trained for camera model identification transfer very well to other forensic tasks such manipulation detection~\cite{mayer2018unified}, suggesting that deep features related to digital forensics are general to a variety of forensics tasks. 

\subsubsection{Architecture}
To build a forensic feature extractor, we adapt the MISLnet CNN architecture developed in \cite{bayar2018mislnet}, which has been utilized in a number of works that target different digital image forensics tasks including manipulation detection~\cite{bayar2016deep,bayar2018mislnet,bayar2017design} and camera model identification~\cite{bayar2017design,mayer2018unified}.
%
Briefly, this CNN consists of 5 convolutional blocks, labeled `conv1' through `conv5' in Fig.~\ref{fig:full_arch} and two fully connected layers labeled `fc1' and `fc2'. Each convolutional block, with the exception of the first, contains a convolutional layer followed by batch normalization, activation, and finally a pooling operation. The two fully connected layers, labeled `fc1' and `fc2,' each consist of 200 neurons with hyperbolic tangent activation. Further details of this CNN are found in \cite{bayar2018mislnet}. 

To use this CNN as a deep feature extractor, an image patch is fed forward through the (trained) CNN. Then, the activated neuron values in the last fully connected layer, `fc2' in Fig.~\ref{fig:full_arch}, are recorded. These recorded neuron values are then used as a feature vector that represents high-level forensic information about the image. The extraction of deep-features from an image patch is the mapping in \eqref{eq:feature_extractor}, where the feature dimension \mbox{$N=200$} corresponding to the number of neurons in `fc2.'\looseness=-1

The architecture of this CNN-based feature extractor is similar to the architecture we used in our prior work in~\cite{mayer2018similarity}. However, in this work we alter the CNN architecture in two ways to improve the robustness of the feature extractor. First, we use full color image patches in RGB as input to the network, instead of just the green color channel used previously. Since many important forensic features are expressed across different color channels, it is important for the network to learn these feature representations. This is done by modifying each 5$\times$5 convolutional kernel to be of dimension 5$\times$5$\times$3, where the last dimension corresponds to image's the color channel. We relax the constraint imposed on the first convolutional layer in \cite{bayar2018mislnet} which is used to encourage the network to learn prediction error residuals. 
This constraint is defined in~\cite{bayar2018mislnet} for single channel image patches, but not color images. Second, we double the number of kernels in the first convolutional layer from 3 to 6, to increase the expressive power of the network.

The feature extractor architecture is depicted by each of the two identical `Feature Extractor' blocks in Fig.~\ref{fig:full_arch}. In our proposed system, we use two identical feature extractors, in `Siamese' configuration~\cite{bromley1994signature}, to map two input image patches $X_1$ and $X_2$ to features $f(X_1)$ and $f(X_2)$. This configuration ensures that the feature extraction step is symmetric, i.e. the ordering of $X_1$ and $X_2$ does not impact the extracted feature values. We refer to the Siamese feature extractor blocks as using \textit{hard sharing}, meaning that the exact same weights and biases are shared between the two blocks.

\subsubsection{Training Methodology}
In our proposed approach we first train the feature extractor during \textit{Learning Phase A}. To do this, we add an additional fully-connected layer with softmax activation to the feature extractor architecture. We provide the feature extractor with image patches and labels associated with the forensic trace of each image patch. Then, the we iteratively train the network using stochastic gradient descent with a cross-entropy loss function. Training is performed for 30 epochs with an initial learning rate of 0.001, which is halved every three epochs, and a batch size of 50 image~patches.

During Learning Phase A we train the feature extractor network on a closed set of forensic traces referred to as ``known forensic traces." Research in~\cite{mayer2018unified} found that training a CNN in this way yields deep-feature representations that are general to other forensic tasks. In particular, it was shown that when a feature extractor was trained for camera model identification, it was very transferable to other forensic tasks. Because of this, during Learning Phase A we train the feature extractor on a large set of image patches with labels associated with their source camera model.

In this work, we train two versions of the feature extractor network: one feature extractor that uses 256$\times$256 image patches as input and another that uses 128$\times$128 image patches as input. We note that to decrease the patch size further would require substantial architecture changes due to the pooling layers. In each case, we train the network using 2,000,000 image patches from the 50 camera models in the ``Camera model set A" found in Table~\ref{tab:camera_models}.

The feature extractor is then updated again in Learning Phase B, as described below in Sec.~\ref{sec:approach:ssec:similarity}. This is significantly different than in our previous work in~\cite{mayer2018similarity}, where the feature extractor remains frozen after Learning Phase A. In our experimental evaluation in Sec.~\ref{sec:experiments1:ssec:cam_model}, we show that allowing the feature extractor to update during Learning Phase B significantly improves system performance.

\subsection{Learning Phase B - Similarity Network}
\label{sec:approach:ssec:similarity}
Here, we describe our proposed neural network architecture that maps a pair of forensic feature vectors $f(X_1)$ and $f(X_2)$ to a similarity score $\in [0,1]$ as described in~\eqref{eq:similarity_map}. The similarity score, when compared to a threshold, indicates whether the pair of image patches $X_1$ and $X_2$ have the same or different forensic traces. We call this proposed neural network the \textit{similarity network}, and is depicted in the right-hand side of Fig.~\ref{fig:full_arch}. Briefly, the network consists of 3 layers of neurons, which we view as a hierarchical mapping of two input features vectors to successive feature spaces and ultimately an output score indicating forensic similarity.

\subsubsection{Architecture}
The first layer of neurons, labeled by `fcA' in Fig.~\ref{fig:full_arch}, contains 2048 neurons with ReLU activation. This layer maps an input feature vector $f(X)$ to a new, intermediate feature space $f_{inter}(X)$. We use two identical `fcA' layers, in Siamese (hard sharing) configuration, to map each of the input vectors $f(X_1)$ and $f(X_2)$ into $f_{inter}(X_1)$ and $f_{inter}(X_2)$. 

This mapping for the kth value of the intermediate feature vector is calculated by an artificial neuron function:
\begin{equation}
f_{k,inter}\left(X\right) = \phi\left( \sum^N_{i=1}w_{k,i}\, f_i\left(X\right) + b_k\right),
\label{eq:artificial_neuron}
\end{equation}
which is the weighted summation, with weights $w_{k,1}$ through $w_{k,N}$, of the $N=200$ elements in the deep-feature vector $f(X)$, bias term $b_k$ and subsequent activation by ReLU function $\phi(\cdot)$. The weights and bias for each element of $f_{inter}(X)$ are arrived at through stochastic gradient descent optimization as described below.

Next the second layer of neurons, labeled by `fcB' in Fig.~\ref{fig:full_arch}, contains 64 neurons with ReLU activation. As input to this layer, we create a vector\vspace{-2mm}
\begin{equation}
f_{concat}(X_1,X_2) = \left[\begin{matrix}
f_{inter}(X_1)\\
f_{inter}(X_2)\\
f_{inter}(X_1) \odot f_{inter}(X_2)
\end{matrix}\right],
\end{equation}
that is the concatenation of $f_{inter}(X_1)$, $f_{inter}(X_2)$ and $f_{inter}(X_1) \odot f_{inter}(X_2)$, where $\odot$ is the element-wise product operation. This layer maps the input vector $f_{concat}(X_1,X_2)$ to a new `similarity' feature space $f_{sim}(X_1,X_2) \in \mathbb{R}^{64}$ using the artificial neuron mapping described in~\eqref{eq:artificial_neuron}. This similarity feature space encodes information about the relative forensic information between patches $X_1$ and $X_2$. 

Finally, a single neuron with sigmoid activation maps the similarity vector $f_{sim}(X_1,X_2)$ to a single score. We call this neuron the `similarity neuron,' since it outputs a single score~$\in [0,1]$, where a small value indicates $X_1$ and $X_2$ contain different forensic traces, and larger values indicate they contain the same forensic trace. In practice, to facilitate training, we use two neurons with softmax and cross-entropy loss, with one neuron indicating ``similar'' and the other indicating ``different.'' During evaluation, we only observe the softmax value of the ``similar'' neuron which behaves like a sigmoid. To make a decision, we compare the similarity score to a threshold $\eta$ typically set to~0.5.

The proposed similarity network architecture differs from our prior work in~\cite{mayer2018similarity} in that we increase the number of neurons in `fcA' from 1024 to 2048, and we add to the concatenation vector the elementwise multiplication of $f_{inter}(X_1)$ and $f_{inter}(X_2)$. Research in~\cite{lee2014speaker} showed that the elementwise product of feature vectors were powerful for speaker verification tasks in machine learning systems. These additions increase the expressive power of the similarity network, and as a result improve system performance.

\subsubsection{Training Methodology}
Here, we describe the second step of the forensic similarity system training procedure, called Learning Phase B. In this learning phase, we train the similarity network to learn a forensic similarity mapping for any type of measurable forensic trace, such as whether two image patches were captured by the same or different camera model, or manipulated by the same or different editing operation. We control which forensic traces are targeted by the system with the choice of training sample and labels provided during training.\looseness=-1

Notably, during Learning Phase B, we allow the error to back propagate through the feature extractor and update the feature extractor weights. This allows the feature extractor to learn better feature representations associated with the type of forensic trace targeted in this learning phase. Allowing the feature extractor to update during Learning Phase B significantly differs from the implementation in~\cite{mayer2018similarity}, which used a frozen feature extractor.

We train the similarity network (and update the feature extractor simultaneously) using stochastic gradient descent for 30 epochs, with an initial learning rate of 0.005 which is halved every three epochs.
The descriptions of training samples and associated labels used in Learning Phase B are described in Sec.~\ref{sec:experiments1}, where we investigate efficacy on different types of forensic traces.
\begin{table*}[t]\centering
\vspace{-4mm}
\caption{\small Camera models used in training (sets A and B) and testing (set C). Note that $A\cap B =  A\cap C = B\cap C = \emptyset$. \newline $^*$Denotes from the Dresden Image Database~\cite{gloe2010dresden}\vspace{-2.5mm}}
\label{tab:camera_models}
\setlength{\tabcolsep}{0.5em}
\scriptsize
\begin{tabular}{l l l l l l l l} 
\toprule
\multicolumn{8}{l}{\textbf{{Camera model set A}}} \\ 
Apple iPhone 4& Agfa Sensor530s$^*$& Canon SX420 IS& LG Nexus 5x& Nikon S710$^*$& Pentax OptioA40$^*$& Samsung L74wide$^*$& Sony NEX-5TL\\ 
Apple iPhone 4s& Canon EOS SL1& Canon SX610 HS& Motorola Maxx& Nikon D200$^*$& Praktica DCZ5.9$^*$& Samsung NV15$^*$& \\ 
Apple iPhone 5& Canon PC1730& Casio EX-Z150$^*$& Motorola Turbo& Nikon D3200& Ricoh GX100$^*$& Sony DSC-H300& \\ 
Apple iPhone 5s& Canon A580& Fujifilm S8600& Motorola X& Nikon D7100& Rollei RCP-7325XS$^*$& Sony DSC-W800& \\ 
Apple iPhone 6& Canon ELPH 160& Huawei Honor 5x& Motorola XT1060& Panasonic DMC-FZ50$^*$& Samsung Note4& Sony DSC-WX350& \\ 
Apple iPhone 6+& Canon S100& LG G2& Nikon S33& Panasonic FZ200& Samsung S2& Sony DSC-H50$^*$& \\ 
Apple iPhone 6s& Canon SX530 HS& LG G3& Nikon S7000& Pentax K-7& Samsung S4& Sony DSC-T77$^*$& \\ 
\midrule
\multicolumn{8}{l}{\textbf{{Camera model set B}}} \\ 
Apple iPad Air 2& Blackberry Leap& Canon SX400 IS& HTC One M7& Motorola Nexus 6& Olympus TG-860& Samsung Note5& Sony A6000\\ 
Apple iPhone 5c& Canon Ixus70$^*$& Canon T4i& Kodak C813$^*$& Nikon D70$^*$& Panasonic TS30& Samsung S3& Sony DSC-W170$^*$\\ 
Agfa DC-733s$^*$& Canon PC1234& Fujifilm XP80& Kodak M1063& Nikon D7000& Pentax OptioW60$^*$& Samsung S5& \\ 
Agfa DC-830i$^*$& Canon G10& Fujifilm J50$^*$& LG Nexus 5& Nokia Lumia 920& Samsung Note3& Samsung S7& \\ 
\midrule
\multicolumn{8}{l}{\textbf{{Camera model set C}}} \\ 
Agfa DC-504$^*$& Canon Ixus55$^*$& Canon Rebel T3i& LG Realm& Nikon D3000& Samsung Lite& Samsung Note2& Sony DSC-T70\\ 
Agfa Sensor505x$^*$& Canon A640$^*$& LG Optimus L90& Nikon S3700& Olympus 1050SW$^*$& Samsung Nexus& Samsung S6 Edge& \\ 
\midrule
\end{tabular}
\vspace{-4mm}
\end{table*}

\vspace{-2.5mm}
\subsection{Patch Selection}
\label{sec:approach:ssec:patch_selection}
Some image patches may not contain sufficient information to be reliably analyzed for forensics purposes~\cite{guera2018reliability}. Here, we describe a method for selecting image patches that are appropriate for forensic analysis. In this paper we use an entropy based selection method to filter out image patches prior to analyzing their forensic similarity. This filter is employed during evaluation only and not while training.

To do this, we view a forensic trace as an amount of information encoded in an image that has been induced by some processing operation. An image patch is a channel that communicates this information. From this channel, we extract forensic information, via the feature extractor, and then compare pairs of these features using the similarity network. Consequently, an image patch must have sufficient capacity in order to encode meaningful forensic information.

When evaluating pairs of image patches, we ensure that both patches have sufficient capacity to encode a forensic trace by measuring their entropy. Here, entropy $h$ is defined as
\vspace{-1mm}%
\begin{equation}
h = -\sum^{255}_{k=0}p_k \ln \left(p_k\right),
\vspace{-1mm}
\end{equation}
where $p_k$ is the probability that a pixel has luminance value $k$ in the image patch. Entropy $h$ is measured in nats. We estimate $p_k$ by measuring the proportion of pixels in an image patch that have luminance value $k$. 

When evaluating image patches, we ensure that both image patches have entropy between 1.8 and 5.2 nats. We chose these values since 95\% of image patches in our database fall within this range. 
Intuitively, the minimum threshold for our patch selection method eliminates flat (e.g. saturated) image patches. Saturated patches have similar appearance regardless of differences in the source camera model or processing history. This method also removes patches with very high entropy. In this case, there is high pixel value variation in the image that may obfuscate the forensic trace. We experimentally validate these threshold choices in Sec.~\ref{sec:experiments1:ssec:cam_model}.
\vspace{-0.5mm}

\vspace{-1mm}
\section{Experimental Evaluation}
\label{sec:experiments1}
We conducted a series of experiments to test the efficacy of our proposed forensic similarity system in different scenarios. In these experiments, we tested our system accuracy in determining whether two image patches were 1)~captured by the same or different camera model, 2)~manipulated by the same or different editing operation, and 3)~manipulated by the same or different manipulation parameter, given a particular editing operation. These scenarios were chosen for their variety in types of forensic traces and because those  traces are targeted in forensic investigations~\cite{bondi2017first,bayar2017augmented,bayar2018mislnet}. Additionally, we conducted experiments that examined properties of the forensic similarity system, including: the effects of patch size and post-compression, comparison to other similarity measures, and the impact of network design and training procedure choices.

The results of these experiments show that our proposed forensic similarity system is highly accurate for comparing a variety of types of forensic traces across two image patches. Importantly, these experiments show this system is accurate even on ``unknown" forensic traces that were not used to train the system. Furthermore, the experiments show that our proposed system significantly improves upon prior art in~\cite{mayer2018similarity}, reducing error rates by over 50\%. 

\begin{figure*}
\vspace{-4mm}
\centering
\includegraphics[width=0.97\linewidth]{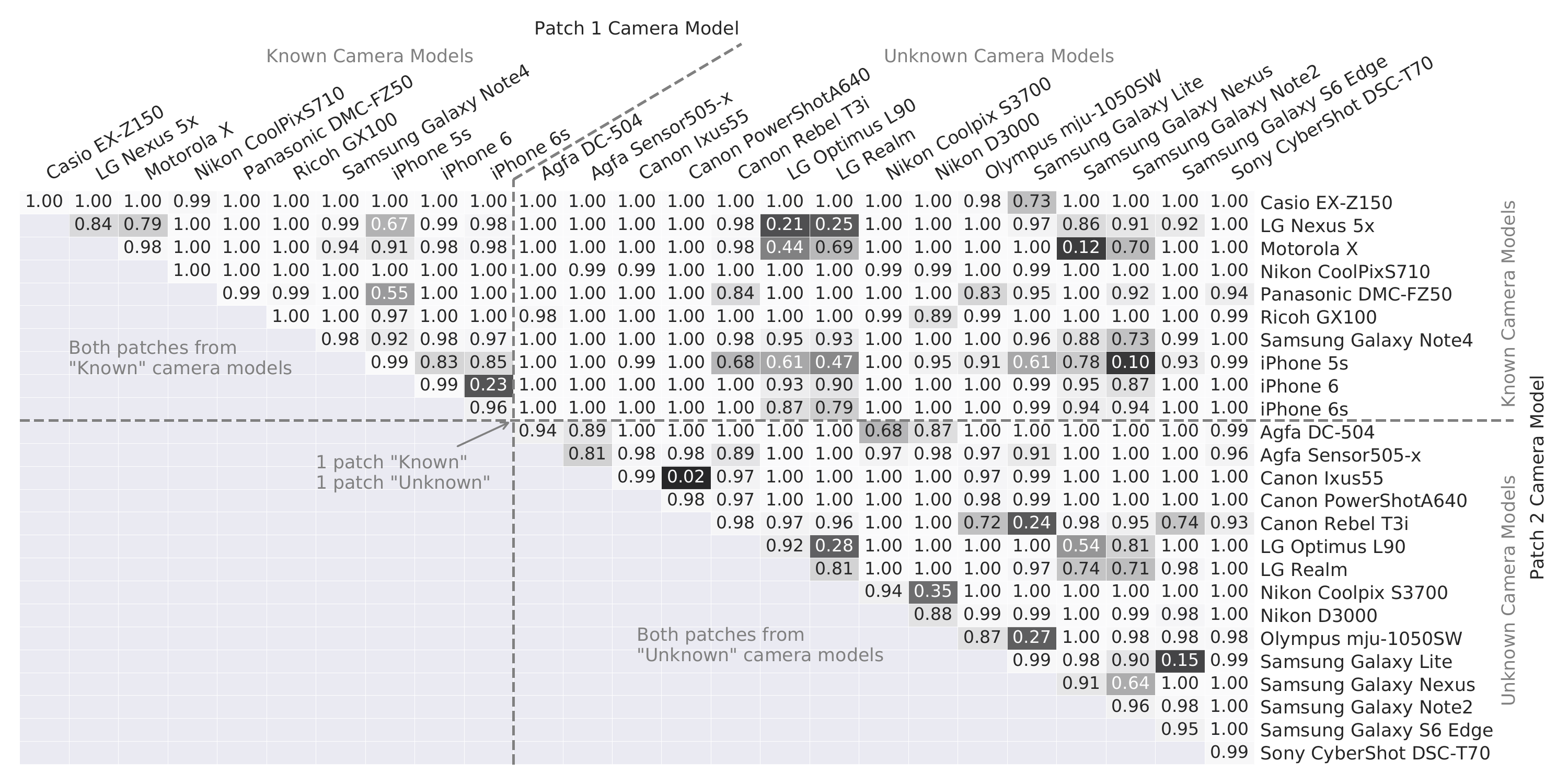}\vspace{-4mm}
\caption{Camera model correct comparison rates for 25 different camera models. Same camera model correct classification rates are on the diagonal, and different camera model correct classification rates are in the non-diagonal entries. Ten camera models were used in training, i.e. ``Known" camera models, and 15 were not used in training, i.e. are ``Unknown" with respect to the classifier. The background color scales with classification rate.}
\label{fig:cam_comparison_mat_256x256}
\vspace{-4mm}
\end{figure*}

To do this, we started with a database of 47,785 images collected from 95 different camera models, which are listed in Table~\ref{tab:camera_models}. Images from 26 camera models were collected as part of the Dresden Image Database ``Natural images" dataset~\cite{gloe2010dresden}. The remaining 69 camera models were from our own database comprised of point-and-shoot, cellphone, and DSLR cameras from which we collected at minimum 300 images with diverse and varied scene content. The camera models were split into three disjoint sets, $\mathcal{A}$, $\mathcal{B}$, and $\mathcal{C}$. Images from $\mathcal{A}$ were used to train the feature extractor in Learning Phase A, images from $\mathcal{A}$ and $\mathcal{B}$ were used to train the similarity network in Learning Phase B, and images from $\mathcal{C}$ were used for evaluation only. First, set $\mathcal{A}$ was determined by randomly selecting 50 camera models from among those that had at least 40,000 non-overlapping 256$\times$256 patches. 
Next, camera model set $\mathcal{B}$ was selected by randomly choosing 30 camera models, from among the remaining, which had at least 25,000 non-overlapping 256$\times$256 patches. Finally, the remaining 15 camera models were assigned to $\mathcal{C}$.

In all experiments, we started with a pre-trained feature extractor that was trained from 2,000,000 randomly chosen image patches from camera models in $\mathcal{A}$ (40,000 patches per model) with labels corresponding to their camera model, as described in Sec.~\ref{sec:approach}. For all experiments, we started with this feature extractor. Research in~\cite{mayer2018unified} showed that deep features related to camera model identification are a good starting point for extracting other types of forensic information, including identification of manipulation operations.

Next, in each experiment we conducted Learning Phase B to target a specific type of forensic trace. To do this, we created a training dataset of pairs of image patches. These pairs were selected by randomly choosing 400,000 image patches of size 256$\times$256 from images in camera model sets $\mathcal{A}$ and $\mathcal{B}$, with 50\% of patch pairs chosen from the same camera model, and 50\% from different camera models. For experiments where the source camera model was compared, a label of 0 or 1 was assigned to each pair corresponding to whether they were captured by different or the same camera model. For experiments where we compared the manipulation type or manipulation parameter, these image patches were then further manipulated (as described in each experiment below) and a label assigned indicating the same or different manipulation type/parameter. Training was performed using Tensorflow v1.10.0 on a Nvidia GTX 1080 Ti.

To evaluate system performance, we created an evaluation dataset of 1,200,000 pairs of image patches, which were selected by randomly choosing 256$\times$256 image patches from the 15 camera models in set $\mathcal{C}$ (``unknown" camera models not used in training). We also included image patches from 10 camera models randomly chosen from set $\mathcal{A}$. One device from each of these 10 ``known" camera models was withheld from training, and only images from these devices were used in this evaluation dataset. For experiments where we compared the manipulation type or manipulation parameter, the pairs of image patches in the evaluation dataset were then further manipulated (as described in each experiment below) and assigned a label indicating the same or different manipulation type/parameter.\looseness=-1

\vspace{-2mm}
\subsection{Source Camera Model Comparison}
\label{sec:experiments1:ssec:cam_model}
In this experiment, we tested the efficacy of our proposed forensic similarity approach for determining whether two image patches were captured by the same or different camera model. To do this, during Learning Phase B we trained
the similarity network using the an expanded training dataset of 1,000,000 pairs of 256$\times$256 image patches selected from camera models in $\mathcal{A}$ and $\mathcal{B}$, with labels indicating whether the source camera model was the same or different. Evaluation was then performed on the evaluation dataset of 1,200,000 pairs from camera models in $\mathcal{A}$ (known) and $\mathcal{C}$ (unknown).\looseness=-1

Fig.~\ref{fig:cam_comparison_mat_256x256} shows the accuracy of our proposed forensic similarity system, broken down by camera model pairing. 
The diagonal entries of the matrix show the correct classification rates of when two image patches were captured by the same camera model. The non-diagonal entries of the matrix show the correct classification rates of when two image patches were captured by different camera models. For example, when both image patches were captured by a Canon Rebel T3i our system correctly identified their source camera model as ``the same" 98\% of the time. When one image patch was captured by a Canon PowerShot A640 and the other image patch was captured by a Nikon CoolPix S710, our system correctly identified that they were captured by different camera models 100\% of the time. 

The overall classification accuracy for all cases was 94.00\%. The upper-left region shows classification accuracy for when two image patches were captured by known camera models, Casio EX-Z150 through iPhone 6s. The total accuracy for the known versus known cases was 95.93\%. The upper-right region shows classification accuracy for when one patch was captured by an unknown camera model, Agfa DC-504 through Sony Cybershot DSC-T70, and the other patch was captured by a known camera model. The total accuracy for the known versus unknown cases was 93.72\%. The lower-right region shows classification accuracy for when both image patches were captured by unknown camera models. For the unknown versus unknown cases, the total accuracy was 92.41\%. This result shows that while the proposed forensic similarity system performs better on known camera models, the system is accurate on image patches captured by unknown camera models.

In the majority of camera model pairs, our proposed forensic similarity system is highly accurate, and achieved $>$95\% accuracy in 257 of the 325 unique pairings of all camera models, and 95 of the 120 possible pairs of unknown camera models. There are also certain pairs where the system does not achieve high comparison accuracy. Many of these cases occurred when two image patches were captured by similar camera models of the same manufacturer. As an example, when one camera model was an iPhone 6 and the other an iPhone 6s, the system only achieved a 26\% correct classification rate. This was likely due to the similarity in hardware and processing pipeline of both of these cellphones, leading to very similar forensic traces. This phenomenon was also observed in the cases of Canon Powershot A640 versus Canon Ixus 55, any combination of LG phones, Samsung Galaxy S6 Edge versus Samsung Galaxy Lite, and Nikon Coolpix S3700 versus Nikon D3000. In only a few cases, low comparison rates were found across camera brands, such as with the Samsung Galaxy Nexus and Motorola X. There are many potential sources of this confusion, such as the licensing of similar technologies, which are the subject of further research. Still, high similarity performance was achieved in the majority of camera model pairings.\looseness=-1

The results of this experiment show that our proposed forensic similarity system is effective at determining whether two image patches were captured by the same or different camera model, even when the camera models were unknown, i.e. not used to train the system. This experiment also shows that, while the system achieves high accuracy in most cases, there are certain pairs of camera models where the system does not achieve high accuracy and this often due to the underlying similarity of the camera model systems themselves.

\subsubsection{Patch Size and Re-Compression Effects}
A forensic investigator may encounter smaller image patches and/or images that have undergone additional compression. 
In this experiment, we examined the performance of our proposed system when presented with input images that have undergone a second JPEG compression and when the patch size is reduced to a of size 128$\times$128.\looseness=-1
\begin{figure}
\centering
\includegraphics[width=0.95\linewidth]{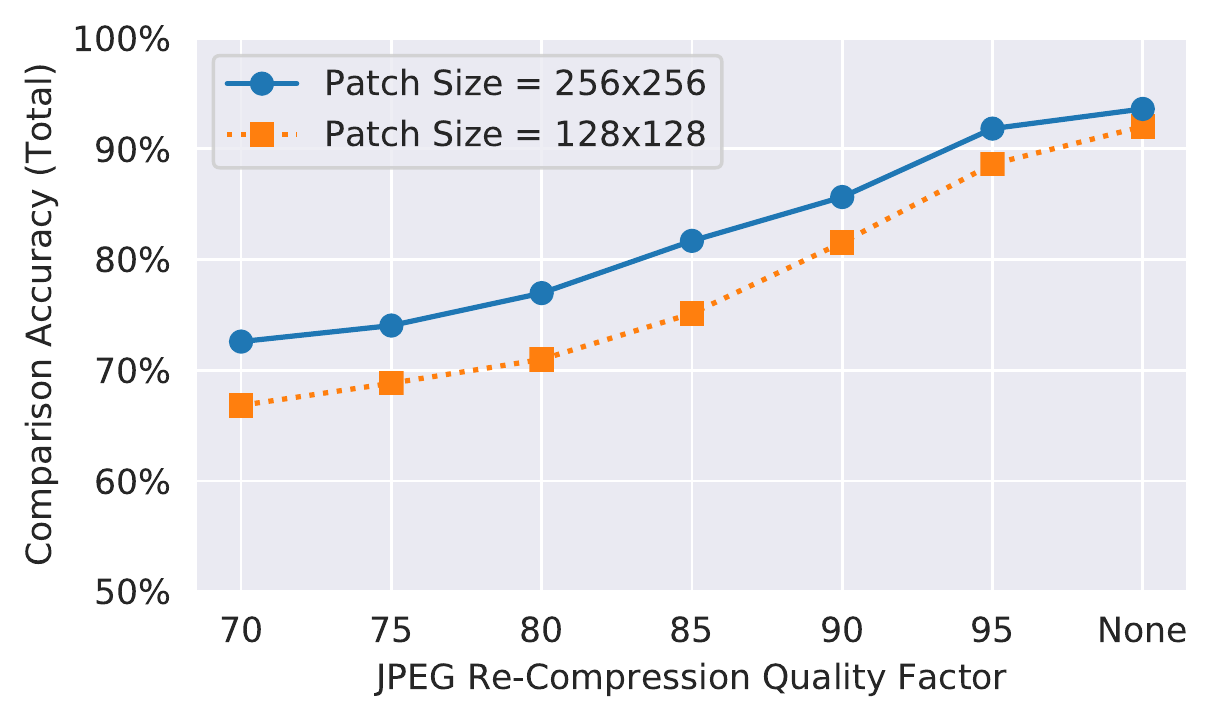}
\vspace{-3mm}
\caption{Camera model comparison accuracy with different patch sizes and JPEG re-compression quality factors.}
\label{fig:jpeg_result}
\vspace{-4mm}
\end{figure}

To do this, we repeated the above source camera model comparison experiment in several scenarios: input patches with size 256$\times$256, input patches of size 128$\times$128, JPEG re-compressed patches of size 256$\times$256, and finally JPEG re-compressed patches of size 128$\times$128. 
We first created copies of the training dataset and evaluation dataset. In these copies, each image in the database was JPEG re-compressed prior to extracting patches. Six such copies were made using JPEG quality factors of 70, 75, 80, 85, 90, and 95.  
We then trained one similarity network (Learning Phase B) per each pairing of patch size and JPEG quality factor. For experiments using 128$\times$128 patches, we used the same 256$\times$256 patches but cropped so only the top-left corner remained.

Fig.~\ref{fig:jpeg_result} shows comparison accuracy at different re-compression quality factors for the two patch sizes. This accuracy includes the ``known versus known'', ``known versus unknown'', and ``unknown versus unknown'' cases. At each quality factor, the network trained with a patch size of 256$\times$256 patch size outperformed the 128$\times$128 patch size. This effect was more pronounced at lower quality factors. For example, at a quality factor of 95, the 256$\times$256 system achieved 91.83\% accuracy whereas the 128$\times$128 system achieved 88.63\% accuracy. At a quality factor of 75, the 256$\times$256 system achieved 74.04\% accuracy whereas the 128$\times$128 system achieved 68.83\% accuracy. The result of this experiment shows that both patch size and JPEG re-compression impact overall system performance. Still, even under moderate re-compression conditions, our system is able to accurately compare forensic traces.

\subsubsection{Other Approaches}
In this experiment, we compared the accuracy of our proposed approach to other approaches including standard distance metrics, support vector machines (SVM), extremely randomized trees (ER Trees), contrastive loss optimization~\cite{hadsell2006dimensionality}, and prior art in~\cite{mayer2018similarity}.\looseness=-1

For the SVM and ER Trees machine learning approaches, we trained each method on deep features of the training dataset extracted by the feature extractor after Learning Phase A. We did this to emulate Learning Phase B where the machine learning approach is used in place of our proposed similarity network. We compared a support vector machine (SVM) with RBF kernel $\gamma=0.01$, $C=1.0$, and an Extremely Randomized Trees (ER Trees) classifier with 800 estimators and minimum split depth of 3. We also compared to the method proposed in~\cite{mayer2018similarity}, and used the same training and evaluation data as with our proposed method.\looseness=-1

\begin{table}[t]\centering 
\caption{Camera model comparison accuracy using standard distances and learned similarity measures}
\label{tab:distance_results}
\vspace{-2mm}
\begin{tabular}{l c c | l r}
\toprule
& \multicolumn{2}{c}{Learning Phase} & \\
\textbf{Distance} & A & B & \textbf{Learned Measure} & \\
\midrule
1 Norm & 92.37\% & 93.06\% & MS`18~\cite{mayer2018similarity} & 85.70\%\\
2 Norm & 92.68\% & 93.28\% & ER Trees & 92.44\%\\
Inf. Norm & 90.98\% & 91.73\% & SVM & 92.84\%\\
Bray-Curtis & 91.98\% & 92.57\% & Contrastive Loss \cite{hadsell2006dimensionality} & 92.99\%\\
Cosine & 92.22\% & 92.87\% & Proposed & \textbf{93.61\%} \\
\bottomrule
\end{tabular}
\vspace{-4mm}
\end{table}

To compare against a popular technique that utilizes a Siamese-based architecture, we compared to the contrastive method proposed in~\cite{hadsell2006dimensionality}. During Learning Phase B, instead of a similarity network, we used the contrastive loss function to further update the feature extractor. This loss function encourages pairs of forensic feature vectors to have large Euclidean distances for different classes, and small Euclidean distances for similar classes.

For the distance measures, we extracted deep features from the evaluation set after Learning Phase A and after Learning Phase B. We measured the distance between each pair of deep features and compared to a threshold. The threshold for each approach was chosen to be the one that maximized total accuracy. For features extracted after Learning Phase A, we used a version of the network which was trained on the 80 camera models in $\mathcal{A}$ and $\mathcal{B}$. This was done to normalize data diversity effects since features extracted after Learning Phase B and learned measures have the benefit of additional training data.\looseness=-1

The total classification accuracy achieved on the evaluation set is shown in Table~\ref{tab:distance_results}, with the proposed system accuracy of 93.61\% shown for reference. For the fixed distance measures, the 2-Norm distance achieved the highest accuracy of 93.28\% using features extracted after Learning Phase B, and 92.68\% using features extracted after Learning Phase A. This result shows that the similarity network improves upon similarity performance over standard distances. The result also shows that Learning Phase B improves the accuracy of standard distances on extracted features. 

For the learned measures, the ER Trees classifier achieved an accuracy of 92.44\%, the SVM achieved an accuracy of 92.84\%, and the contrastive loss function achieved an accuracy of 92.99\%, all lower than the proposed similarity network. We also compared against the architecture and training procedure proposed in our previous work~\cite{mayer2018similarity}, which achieved a total accuracy of 85.70\%. The results of this experiment show that our proposed system outperforms other distance measures and learned similarity measures, and significantly improves upon prior work in~\cite{mayer2018similarity} decreasing the comparison error rate by over 50\%.\looseness=-1

\subsubsection{Impact of Training Procedure} 
In this experiment, we examined the effects of two design aspects in the Learning Phase B training procedure. In particular, these aspects are 1) allowing the feature extractor to update, i.e. \textit{unfrozen} during training, and 2) using a diverse training dataset. This experiment was conducted to explicitly compare to the training procedure in~\cite{mayer2018similarity}, where the feature extractor was not updated (\textit{frozen}) in Learning Phase B and only a subset of available training camera models were used.

To do this, we created an additional training database of 400,000 image patch pairs of size 256$\times$256, mimicking the original training dataset, but containing only image patches captured by camera models in set $\mathcal{B}$. This was done since the procedure in~\cite{mayer2018similarity} specified to conduct Learning Phase B on camera models that were not used in Learning Phase A. We refer to this as training set B, and the original training set as AB. We then performed Learning Phase B using each of these datasets. Furthermore, we repeated each training scenario where the learning rate multiplier in each layer in the feature extractor layer was set to 0, i.e. the feature extractor was \textit{frozen}. This was done to compare to the procedure in~\cite{mayer2018similarity} which used a frozen feature extractor.
\begin{table}[t]\centering 
\vspace{-2mm}
\caption{Performance of different training methods}
\label{tab:training_methods_results}
\vspace{-2mm}
\begin{tabular}{l l r}
\toprule
Training Data &  Feature Extractor & Accuracy\\
\midrule
B & Frozen & 90.24\%\\
B & Unfrozen & 90.96\%\\
AB & Frozen & 92.56\%\\
AB & Unfrozen & \textbf{93.61\%}\\
\bottomrule
\end{tabular}
\vspace{-4mm}
\end{table}

The overall accuracy achieved by each of the four scenarios is shown in Table~\ref{tab:training_methods_results}. 
When using training on set B with a frozen feature extractor, which is the same procedure used in~\cite{mayer2018similarity}, the total accuracy on the evaluation image patches was 90.24\%. When allowing the feature extractor to update, accuracy increased by 0.72 percentage points to 90.96\%. When increasing training data diversity to camera model set AB, but using a frozen feature extractor the accuracy achieved was 92.56\%. Finally, when using a diverse dataset and an unfrozen feature extractor, total accuracy achieved was 93.61\%. 

The results of this experiment show that our proposed training procedure is a significant improvement over the procedure using in~\cite{mayer2018similarity}, improving accuracy 3.37 percentage points. Furthermore, we can see the added benefit of our proposed architecture enhancements when comparing the result MS`18 in Table~\ref{tab:distance_results}, which uses both the training procedure and system architecture of~\cite{mayer2018similarity}. Improving the system architecture alone raised classification rates from 85.70\% to 90.24\%. Improving the training procedure further raised classification rates to 93.61\%, together reducing the error rate by more than half.

\subsubsection{Architecture Variants}
In this experiment, we examined the impact of design choices related to the proposed feature extractor and similarity network. To do this, we repeated the initial experiment with several variations on system architecture.
\looseness=-1

Table~\ref{tab:arch_diff_results} shows the overall camera model comparison accuracy using different architecture variants. In the first case, we used only the green channel as input. In this case, 91.13\% accuracy was achieved, increasing the error rate by 2.48 percentage points. This equates to a 39\% relative increase in error from the 93.61\% accuracy achieved by the proposed system. The \textit{relative error increase} (REI) is calculated by the formula $REI=\tfrac{\left(Acc_1-Acc_2\right)}{\left(100-Acc_1\right)}$, where $Acc_1$ is the accuracy achieved the proposed system, and $Acc_2$ is the accuracy achieved by the variations of the proposed architecture. The REI captures the percent change in error rate by using a variant of the proposed architecture.

In the second case, we also used the green channel as input and a convolutional constraint on the first layer according to \cite{bayar2018mislnet}. This was the configuration used in our previous work in~\cite{mayer2018similarity}. In this case, 91.45\% accuracy was achieved. This result shows that the proposed full color information is important for camera model comparisons.

In a third case, we removed the elementwise-multiplication structure in the similarity network. In this case, the error rate increased by 0.51 percentage points, or an 8\% relative error increase. In the final case, we removed the entropy-based patch filter, which increased the error rate by 0.59 percentage points, or a 9\% relative error increase.

The results of this experiment show that architecture improvements over prior work~\cite{mayer2018similarity} led to improved forensic similarity performance. Notably, moving to full color input from green-channel-only led to most significant improvement.

\begin{table}
\vspace{-2mm}
\centering 
\caption{Camera Model Comparison Performance by Architecture Variant, and Associated Relative Error Increase (REI)}
\label{tab:arch_diff_results}
\vspace{-2mm}
\begin{tabular}{l r r r}
\toprule
System Attribute & Accuracy & Difference & REI\\
\midrule
\textbf{Proposed} & \textbf{93.61\%} & \\
Green channel only & 91.13\% & -2.48 & 39\%\\
Green channel only, with constraint & 91.45\% & -2.16 & 34\%\\
Without elementwise multiplication & 93.10\% & -0.51 & 8\%\\
Without entropy filtering &  93.02\% & -0.59 & 9\% \\
\bottomrule
\end{tabular}
\vspace{-4mm}
\end{table}

\subsubsection{Impact of Entropy Threshold}
In this experiment, we investigated the impact of patch filtering using entropy thresholds on performance of forensic similarity. To do this, we repeated the initial camera model comparison experiment and varied the minimum and maximum entropy thresholds described in Sec.~\ref{sec:approach:ssec:patch_selection}.

\begin{figure}
\centering
\captionsetup[subfloat]{captionskip=0mm}

\subfloat[Accuracy by Entropy Threshold]{\includegraphics[height=1.55in]{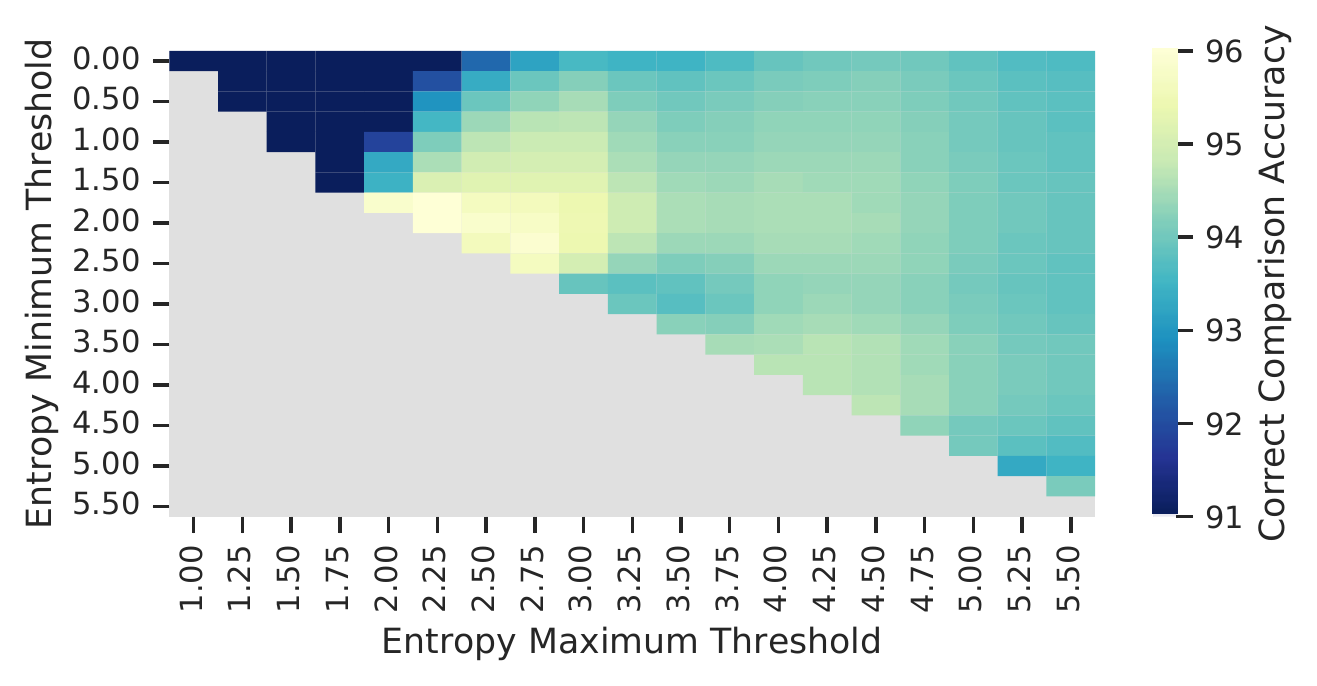}}\vspace{-2mm}

\subfloat[Patch Distribution by Entropy]{\hspace{-8mm}\includegraphics[height=1.25in]{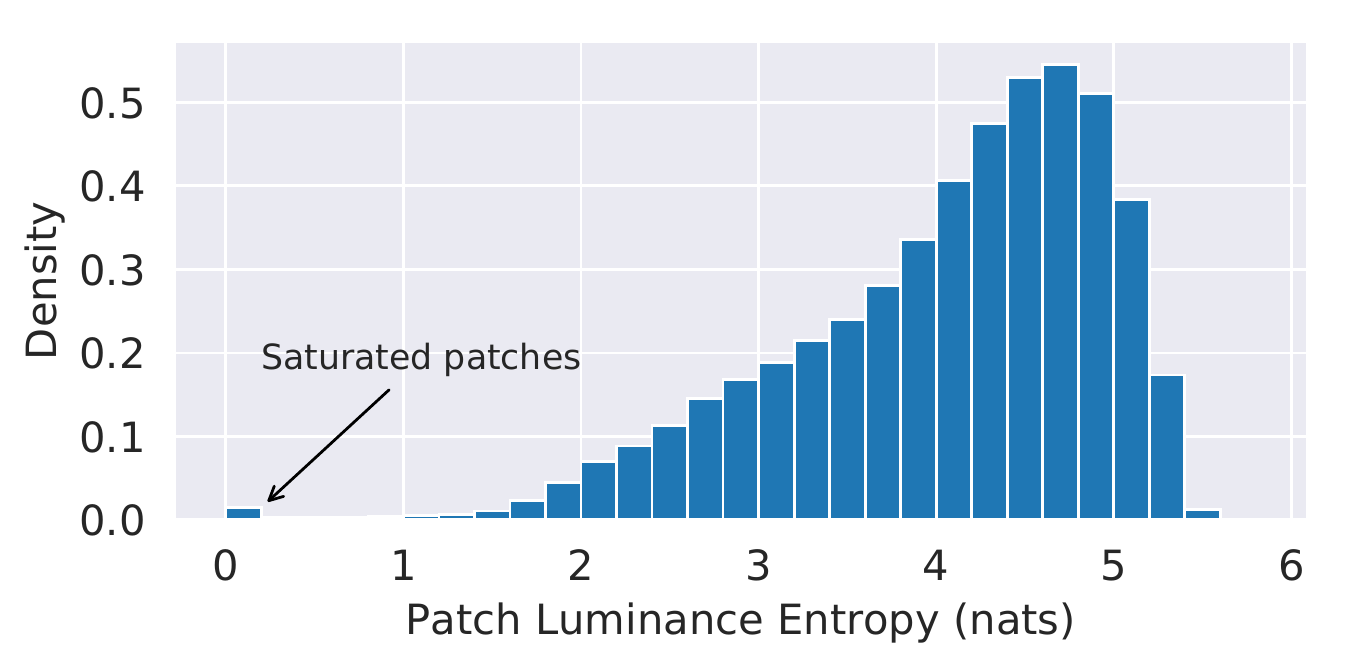}}

\caption{(a) Camera model comparison accuracy at different entropy thresholds, and (b) the distribution of patches by entropy in our testing dataset. The patch entropy has significant impact on forensic similarity performance.}
\label{fig:entropy_test}
\vspace{-4mm}
\end{figure}

Fig.~\ref{fig:entropy_test}(a) shows the overall correct comparison rate at different entropy thresholds. When both patches have low entropy between 0 and 1.75, the comparison accuracy is 78.71\%. These patches are typically saturated, where it is difficult to extract meaningful forensic information. When patches have entropy between 2 and 2.75, the overall comparison accuracy is significantly higher at 95.73\%. These patches are typically ``flat" in appearance, but not saturated. Our intuition for why these patches work well is that the system is able to more effectively separate forensic traces from relatively uniform scene content. When patches have high entropy between 5 and 5.4, the overall comparison accuracy is lower at 93.45\%. Our intuition for why these patches are slightly harder to compare is that they contain highly varying scene content, which obfuscates the forensic traces. These intuitions are corroborated by findings in~\cite{guera2018reliability}, which found that patches with high semantic content were often less reliable for forensics purposes.\looseness=-1

Different applications have different tolerances for the amount of patches needed for analysis. Fig.~\ref{fig:entropy_test}(b) shows the distribution of entropy calculated from patches in our testing dataset. We chose a minimum entropy of 1.8 and maximum of 5.2 since they are relatively permissive, allowing for 95\% of patches to be analyzed while maximizing performance. In applications that can be more selective with patches, higher performance can be achieved with greater selectivity of patches.\looseness=-1

\subsubsection{Unknown Brands}
In the above experiments, we tested forensic similarity performance on unknown camera models. The unknown camera models in Camera Model Set C were captured by brands, such as Apple or Canon, that were also within the training sets. In this experiment, we test performance on brands that do not exist within the training set. To do this, we used the VISION database~\cite{shullani2017vision} and selected 6 camera models with brands not in Camera Model Sets A or B. These camera models are: Asus ZenFone 2 Laser, Lenovo P70A, Microsoft Lumia640LTE, OnePlus A3000, Wiko Ridge4G, and Xiaomi RedmiNote3. Then, we randomly selected 1000 pairs of patches from each pairing of camera models from the ``natural scene'' images within this database, and calculated the forensic similarity between these pairs of patches.

\begin{figure}
\centering
\includegraphics[width=0.755\linewidth]{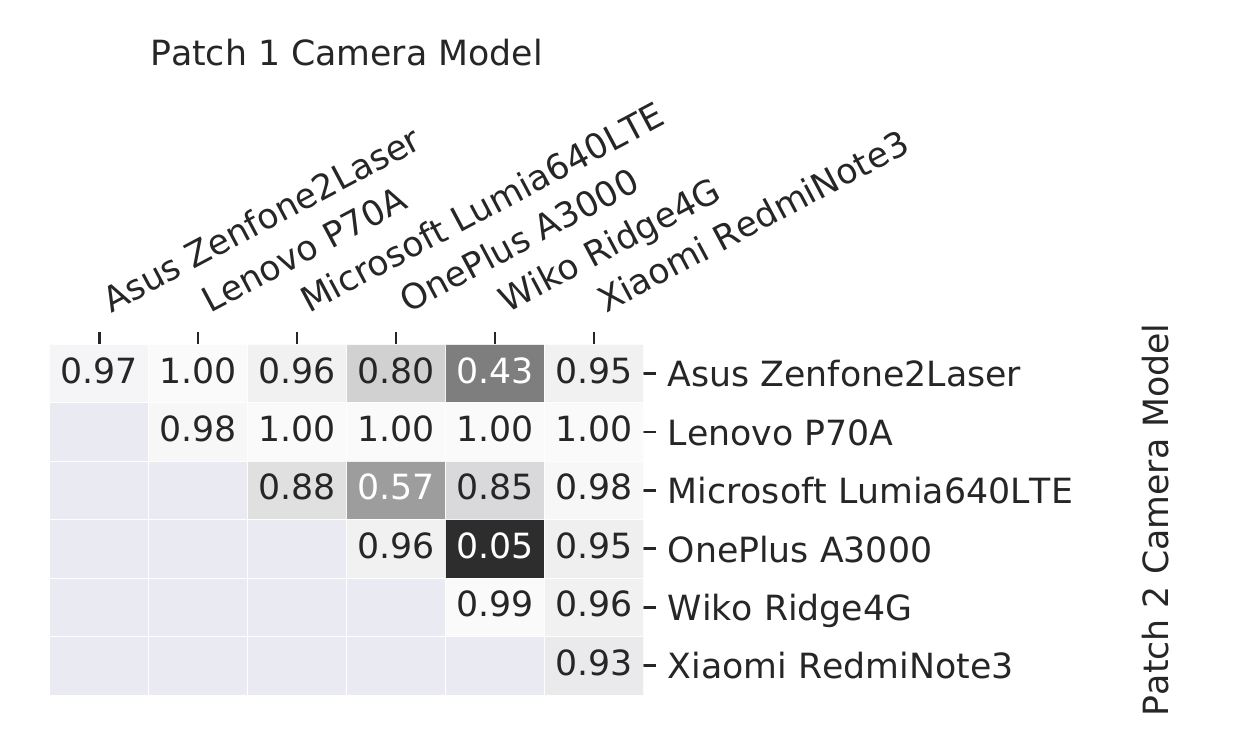}
\vspace{-4mm}
\caption{Camera model comparison rates for camera brands not used in training.}
\label{fig:unknown_brand_test}
\vspace{-4mm}
\end{figure}

Correct comparison rates for these unknown brands are shown in Fig.~\ref{fig:unknown_brand_test}. When both patches were captured by the same camera model, four out of the six camera models achieved 95\% or greater correct comparison accuracy. When patches were captured by different camera models, ten out of the fifteen pairings achieved 95\% or greater correct comparison accuracy. The overall comparison accuracy achieved was 89.77\%, which is slightly less than the 92.41\% accuracy achieved for the original experiment with unknown camera models but with known brands. Forensic similarity performance was high for some camera models in this experiment, such as with the Lenovo P70A which achieved 98\% or greater correct comparison rates in all cases. The approach did not perform as well in other cases, such as when comparing patches from a OnePlus A3000 camera model and WikoRidge 4G camera model, which was only able to correctly differentiate between 5\% of patch pairs. The exact mechanisms or features that were confused between these camera models is not known, and is a topic for further investigation.

The result of this experiment shows that forensic similarity comparisons are effective even on brands of camera models that were not used to train the system. High comparison accuracy was achieved in the majority of camera model pairings.
\looseness=-1

\vspace{-2mm}
\subsection{Editing Operation Comparison}

A forensic investigator is often interested in determining whether two image patches have the same processing history. 
In this experiment, we investigated the efficacy of our proposed approach for determining whether two image patches were manipulated by the same or different editing operation, including ``unknown" editing operations not used to train the system.\looseness=-1 

To do this, we started with the training database of image patch pairs. We then modified each patch with one of the eight ``known" manipulations in Table~\ref{tab:exp_manips}, with a randomly chosen editing parameter. We manipulated 50\% of the image patch pairs with the same editing operation, but with different parameter, and manipulated 50\% of the pairs with different editing operations. The known manipulations were the same manipulations used in~\cite{bayar2016deep} and~\cite{boroumand2018deep}. We repeated this for the evaluation database, using both the ``known" and ``unknown" manipulations. Wiener filtering was performed using the SciPy python library, web dithering was performed using the Python Image Library, and salt and pepper noise was performed using the SciPy image processing toolbox (skimage). We note that the histogram equalization and JPEG compression manipulations were performed on the whole image. We then performed Learning Phase B using the manipulated training database, with labels associated with each  pair corresponding to whether they have been manipulated by the same or different editing operation. Finally, we evaluated accuracy on the evaluation dataset, with patches processed in a similar manner.

\begin{figure}
\vspace{-0mm}
\centering
\includegraphics[width=\linewidth]{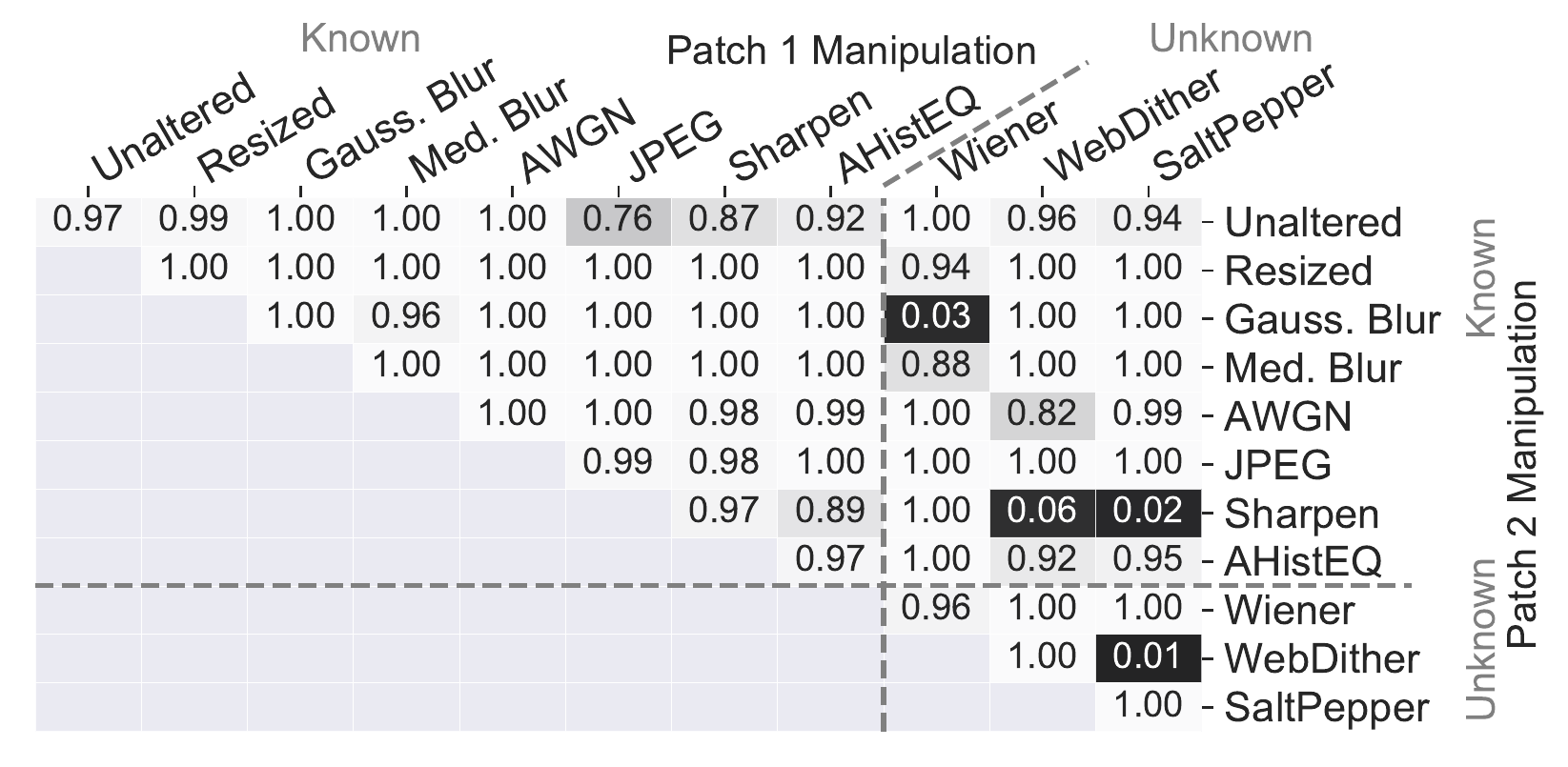}
\vspace{-6.66mm}
\caption{Correct comparison rates when comparing manipulation type of two image patches, with known and unknown manipulations.}
\vspace{-2mm}
\label{tab:manip_results}
\end{figure}

Fig.~\ref{tab:manip_results} shows the correct classification rates of our proposed forensic similarity system, broken down by manipulation pairing, with the first eight columns and rows corresponding to known manipulations, and the final three corresponding to the unknown manipulations. For example, when one image patch was manipulated with salt and pepper noise and the other patch was manipulated with histogram equalization, our proposed system correctly identified that the two patches were edited by different manipulations at a rate of 95\%. When both image patches were edited with Wiener filtering, the system correctly identified that they were edited by the same manipulation at a rate of 96\%.\looseness=-1

The system achieved high comparison accuracy in the vast majority of cases. However, there were certain pairs of manipulations which proved difficult to correctly compare. One critical case occurred when one patch was unaltered and one patch was JPEG recompressed. This scenario had a 76\% correct comparison rate. This result is not unexpected, as all unaltered patches are from images naturally in JPEG format. Research in~\cite{barni2017aligned} showed that there are many cases in which double JPEG compression cannot be differentiated from single JPEG compression, including when the original JPEG quality factor is similar to or greater than the second JPEG quality factor. Such cases are often encountered in the evaluation set. Other difficult cases occurred when one patch was unaltered and the other was sharpened or histogram equalized. A potential explanation for this is that sharpening and histogram equalization induce only slight forensic traces in patches where no edge content exists or when the histogram is already equalized.\looseness=-1

\begin{table}
\caption{Known manipulations used in training and unknown manipulations used in evaluation, with associated parameters}
\begin{center}
\label{tab:exp_manips}
\begin{tabular}{l l l} 
\toprule
Manipulation & Parameter & Value Range\\
\midrule
\multicolumn{3}{l}{\textbf{Known Manipulations}} \\ 
Unaltered & $-$ & $-$   \\
Resizing (bilinear) & Scaling factor & $[0.6,0.9] \cup [1.1,1.9]$\\
Gaussian blur (5$\times$5) & $\sigma$ & $[1.0, 2.0]$ \\
Median blur & Kernel size & $\left\lbrace3,5,7\right\rbrace$\\
AWG Noise & $\sigma$ & $[1.5, 2.5]$ \\
JPEG Compression & Quality factor & $\left\lbrace50, 51, 52,\ldots, 95\right\rbrace$ \\
Unsharp mask $(r=2, t=3)$ & Percent & $[50, 200]$ \\
Adaptive Hist. Eq. & $-$  & $-$ \\
\midrule
\multicolumn{3}{l}{\textbf{Unknown Manipulations}} \\ 
Weiner filter & Kernel size & $\left\lbrace3,5,7\right\rbrace$\\
Web dithering & $-$  &  $-$ \\
Salt + pepper noise & Percent & $\left\lbrace5, 6, 7,\ldots, 20\right\rbrace$ \\
\bottomrule
\end{tabular}
\end{center}
\vspace{-4mm}
\end{table}

The majority of scenarios involving unknown manipulations similarly achieved high comparison accuracy. Correct comparison accuracy of 90\% or greater was achieved in 19 of 24 unknown versus known cases, and in 5 of 6 unknown versus unknown cases. However, some unknown manipulations were difficult to correctly compare. The scenario of Wiener filtering versus Gaussian blurring incorrect comparisons were likely due to the smoothing similarities between Wiener filtering and Gaussian blurring. Web dithering versus sharpening, salt and pepper versus sharpening, and web dithering versus salt and pepper noise also were challenging. The cases are likely due to the addition of similar high frequency artifacts introduced by these operations. When including these unknown manipulations in training, correct comparison accuracy over 98\% is achieved in these cases. This highlights the need for a breadth of training manipulations for a well generalized system.

The results of this experiment demonstrate that our proposed forensic similarity is system is effective at comparing the processing history of image patches, even when image patches have undergone an editing operation that was unknown, i.e. not used during training.

\vspace{-2mm}
\subsection{Editing Parameter Comparison}
\begin{figure}
\centering
\includegraphics[width=0.95\linewidth]{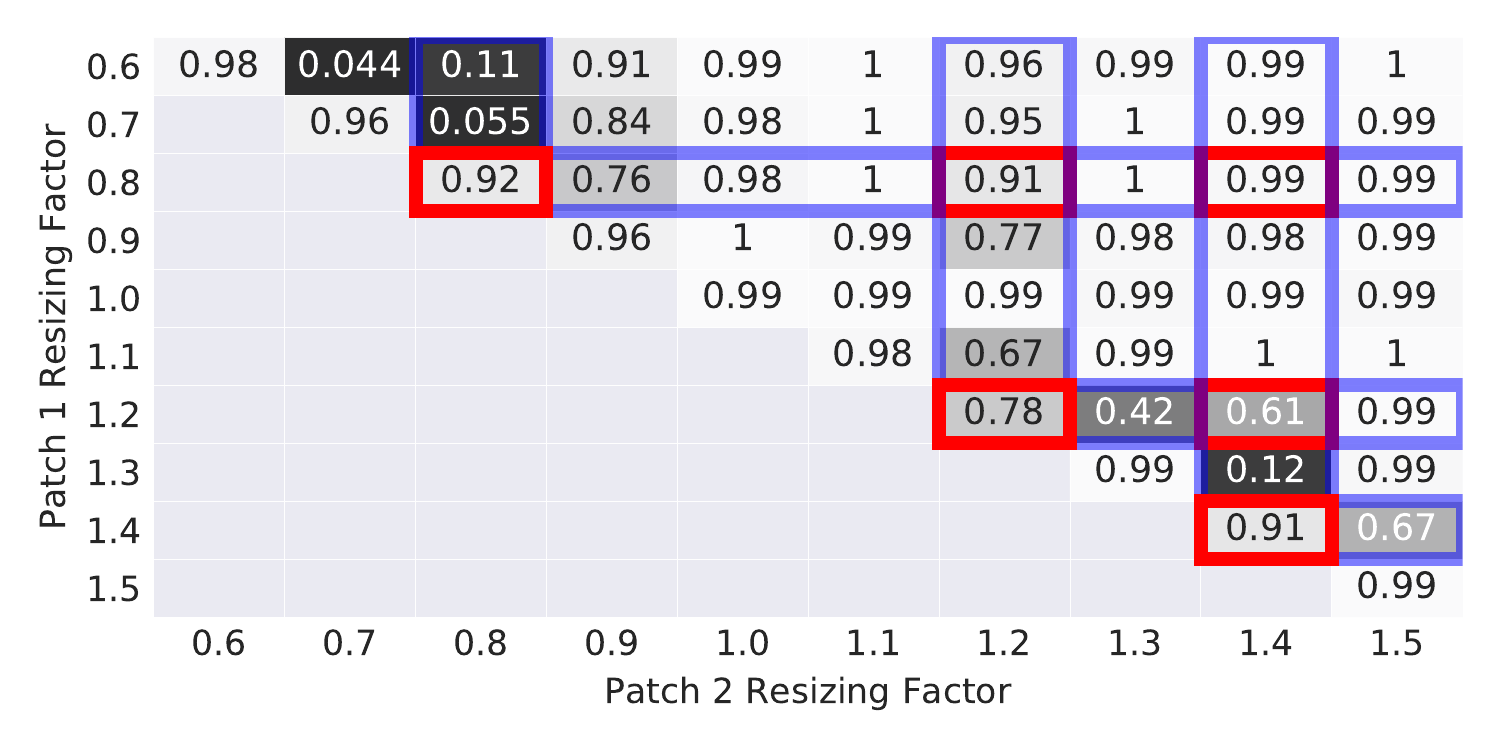}\vspace{-5mm}
\caption{
Correct comparison rates for comparing the resizing parameter in two image patches. Unknown scaling factors are $\lbrace 0.8, 1.2, 1.4\rbrace$. %
Blue highlights one patch with unknown scaling factor, red highlights both patches have unknown scaling factors.
}
\label{fig:param_results}
\vspace{-4mm}
\end{figure}

In this experiment, we investigated the efficacy of our proposed approach for determining whether two image patches have been manipulated by the same or different manipulation parameter. Specifically, we examined pairs of image patches that had been resized by the same scaling factor or that had been resized by different scaling factors, including ``unknown" scaling factors that were not used during training. This type of analysis is important when analyzing spliced images where both the host image and foreign content were resized, but the foreign content was resized by a different factor.

To do this, we started with the training database of image patch pairs. We then resized each patch with one of the seven ``known" resizing factors in $\lbrace 0.6, 0.7, 0.9, None, 1.1, 1.3, 1.5\rbrace$ using bilinear interpolation. We resized 50\% of the image patch pairs with the same scaling factor, and resized 50\% of the pairs with different scaling factors.  We repeated this for the evaluation database, using both the ``known" scaling factors and ``unknown" scaling factors in $\lbrace 0.8, 1.2, 1.4\rbrace$. We then performed Learning Phase B using the training database of resized image patches, with labels corresponding to whether each pair of image patches was resized by the same or different scaling factor.

The correct classification rates of our proposed approach are shown in Fig.~\ref{fig:param_results}, broken down by tested resizing factor pairings. For example, when one image patch was resized by a factor of 0.8 and the other image patch was resized by a factor of 1.4, both unknown scaling factors, our proposed system correctly identified that the image patches were resized by different scaling factors at rate of 99\%. Cases where at least one patch has been resized with an unknown scaling factor are highlighted in blue. Cases where both patches have been resized with an unknown scaling factor our outlined in red. 

Our system achieved greater than 90\% correct classification rates in 33 of 45 tested scaling factor pairings. There are also some cases where our proposed system does not achieve high accuracy. These cases tend to occur when presented with image patches that have been resized with different but similar resizing factors. For example, when resizing factors of 1.4 and 1.3 are used, the system correctly identifies the scaling factor as different 12\% of the time.

The results of this experiment show that our proposed approach is effective at comparing the manipulation parameter in two image patches, a third type of forensic trace. This experiment shows that our proposed approach is effective even when one or both image patches have been manipulated by an unknown parameter of the editing operation not used in training.\looseness=-1

\vspace{-2mm}
\section{Practical applications}
\label{sec:experiments2}
The forensic similarity approach is a versatile technique that is useful in many different practical applications. In this section, we demonstrate how a forensic similarity is used in two types of forensic investigations: image forgery detection and localization, and image database consistency verification.

\vspace{-2mm}
\subsection{Forgery Detection and Localization}
\label{sec:splicing_experiments}

\begin{figure*}
\newcommand{\ww}{1.34in}
\centering
\null\hfill
\subfloat[Original]{\includegraphics[width=1.275in]{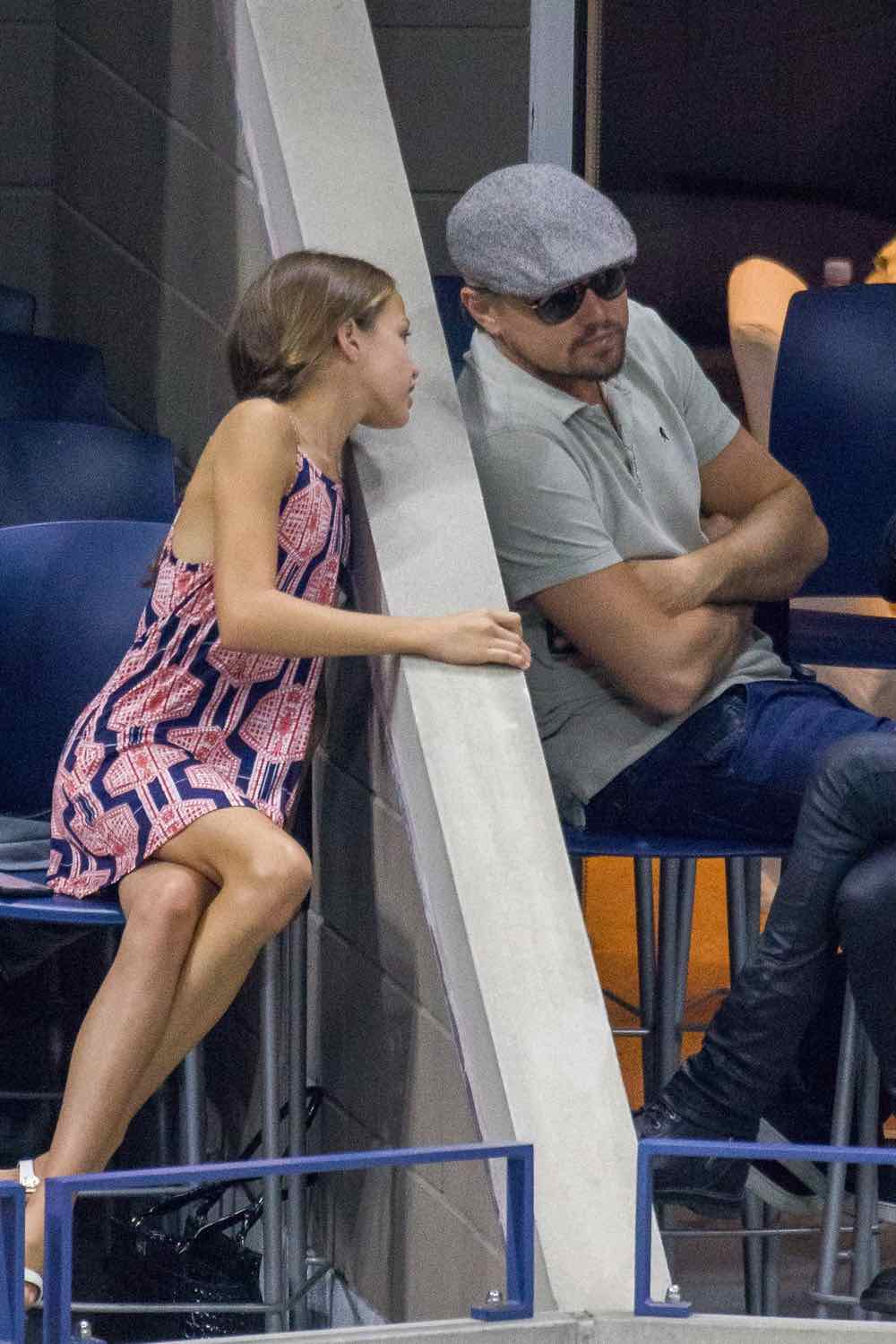}%
\label{fig_first_case}}
\hfill
\subfloat[Spliced]{\includegraphics[width=1.275in]{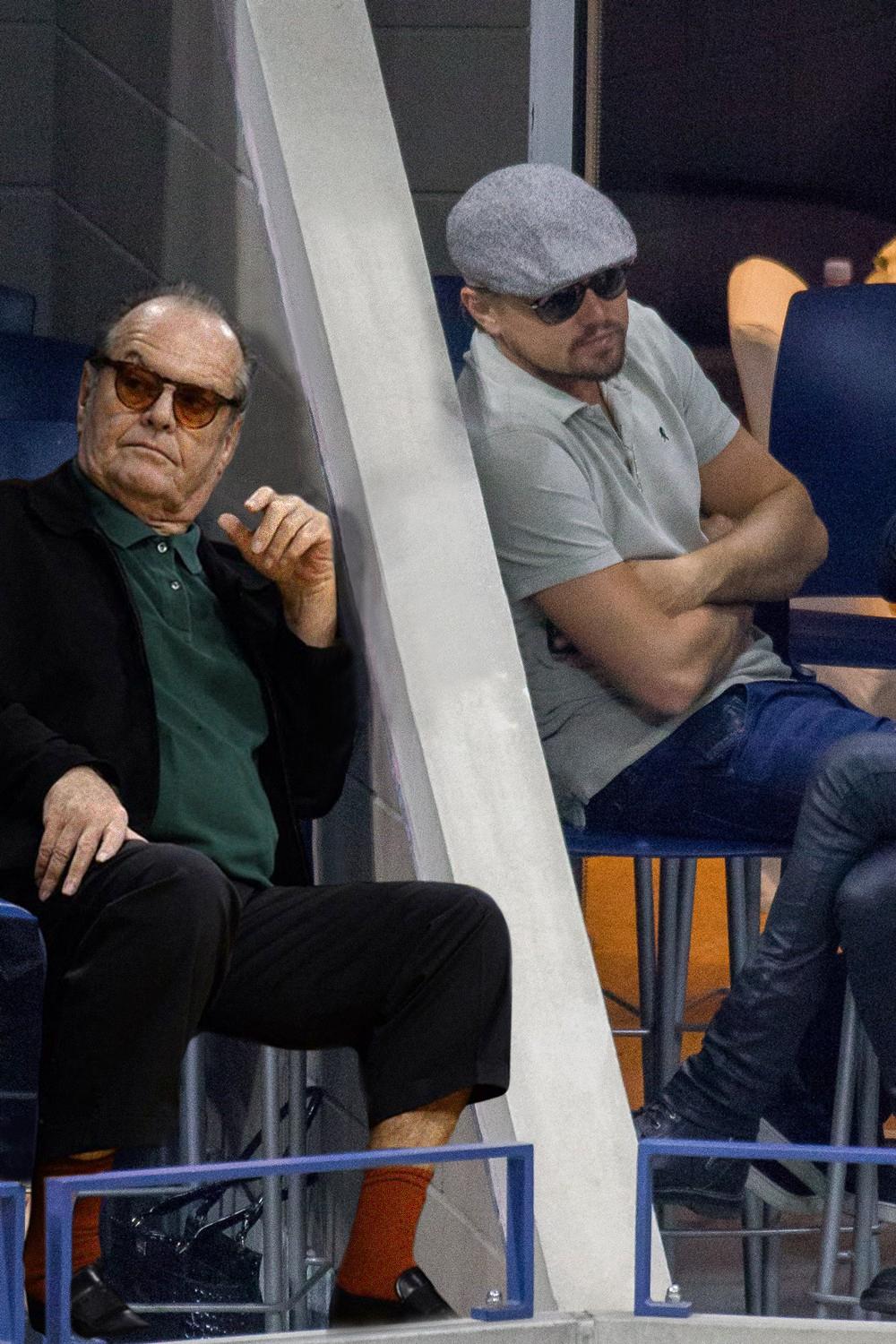}%
\label{fig_second_case}}
\hfill
\subfloat[Host reference]{\includegraphics[width=\ww]{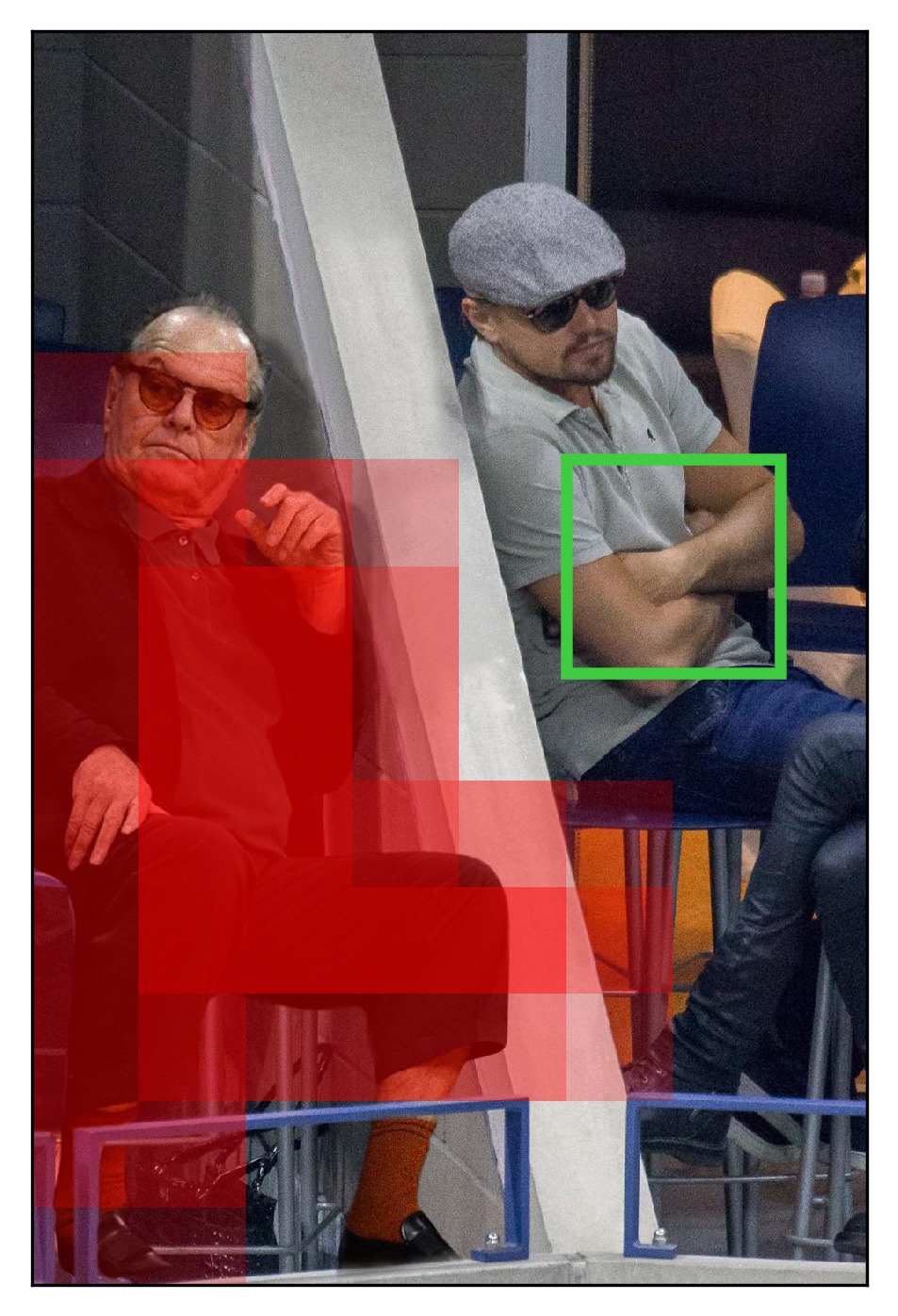}%
}
\hfill
\subfloat[Spliced reference]{\includegraphics[width=\ww]{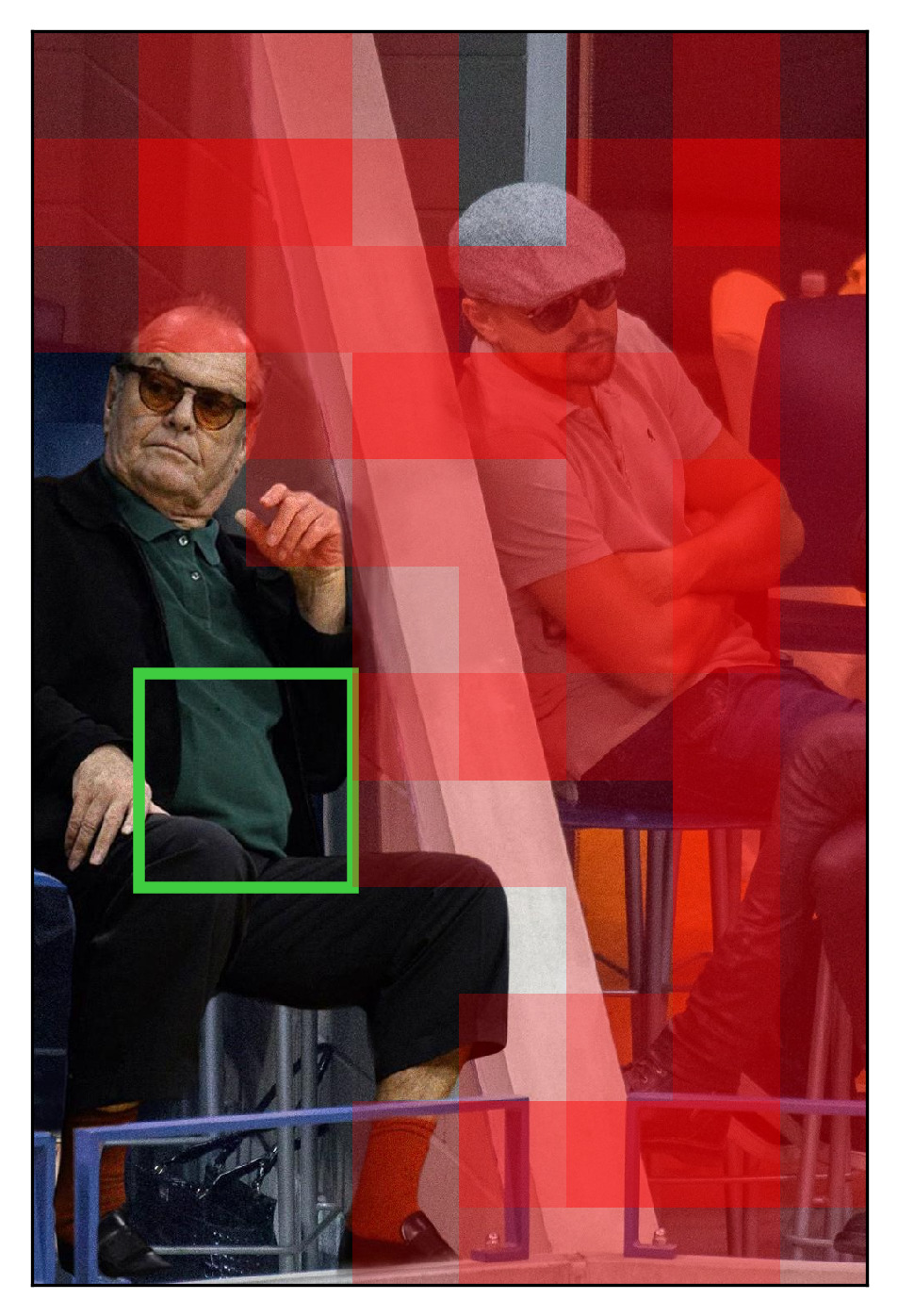}%
}
\hfill
\subfloat[Host ref, original image]{\includegraphics[width=\ww]{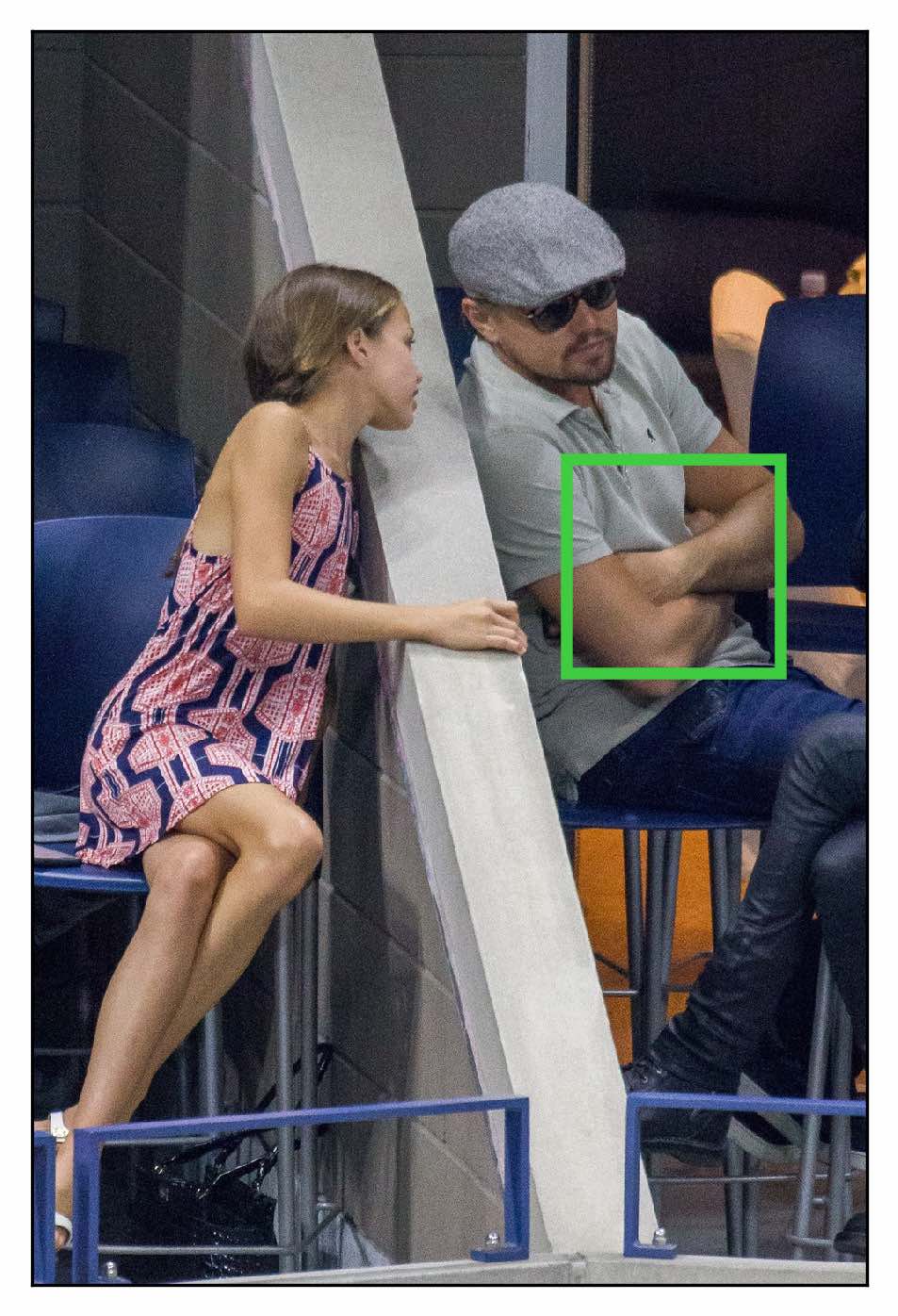}%
}
\hfill\null

\null\hfill
\subfloat[Cam. traces 256x256]{\includegraphics[width=\ww]{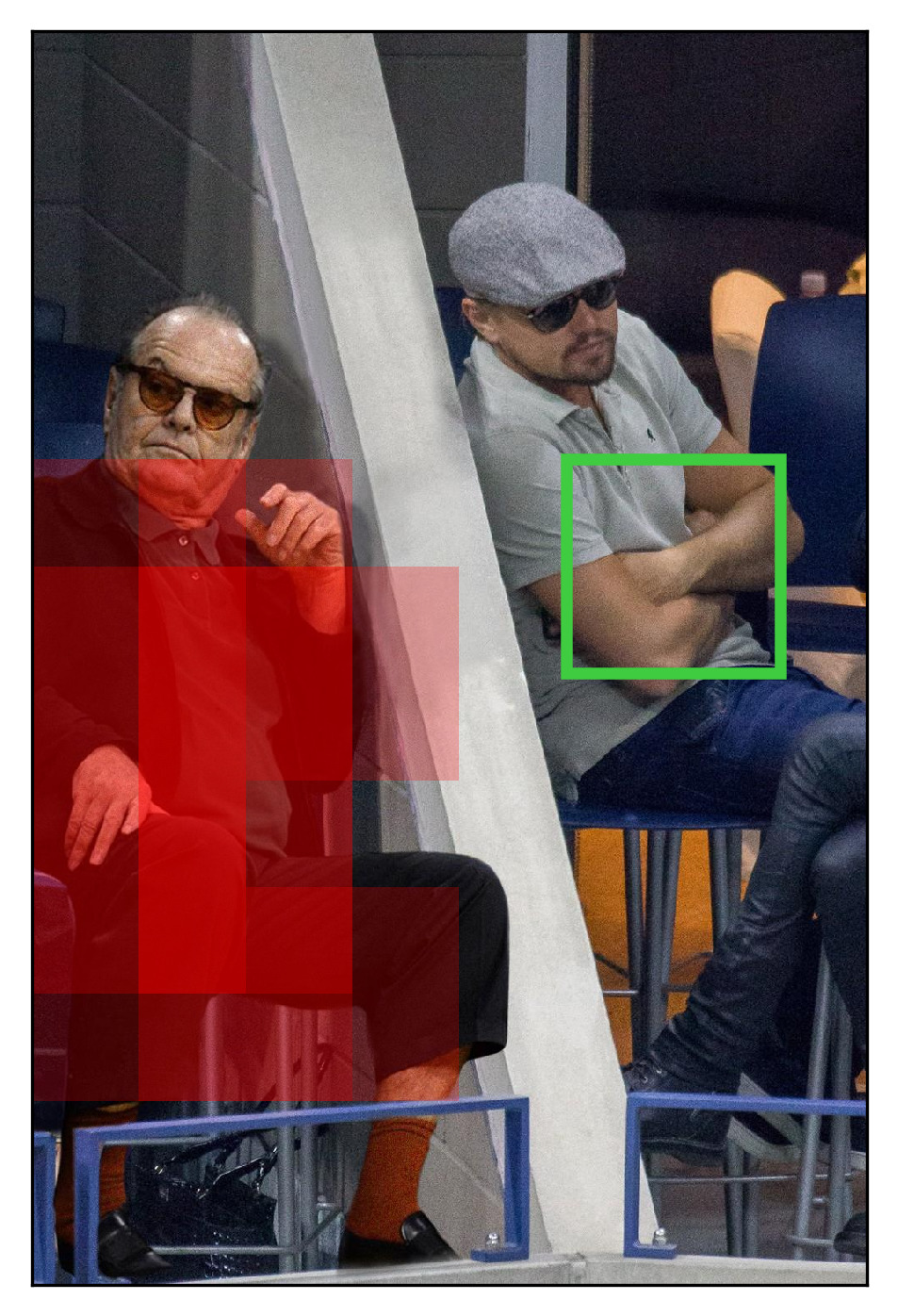}%
}
\hfill
\subfloat[Cam. traces 128x128]{\includegraphics[width=\ww]{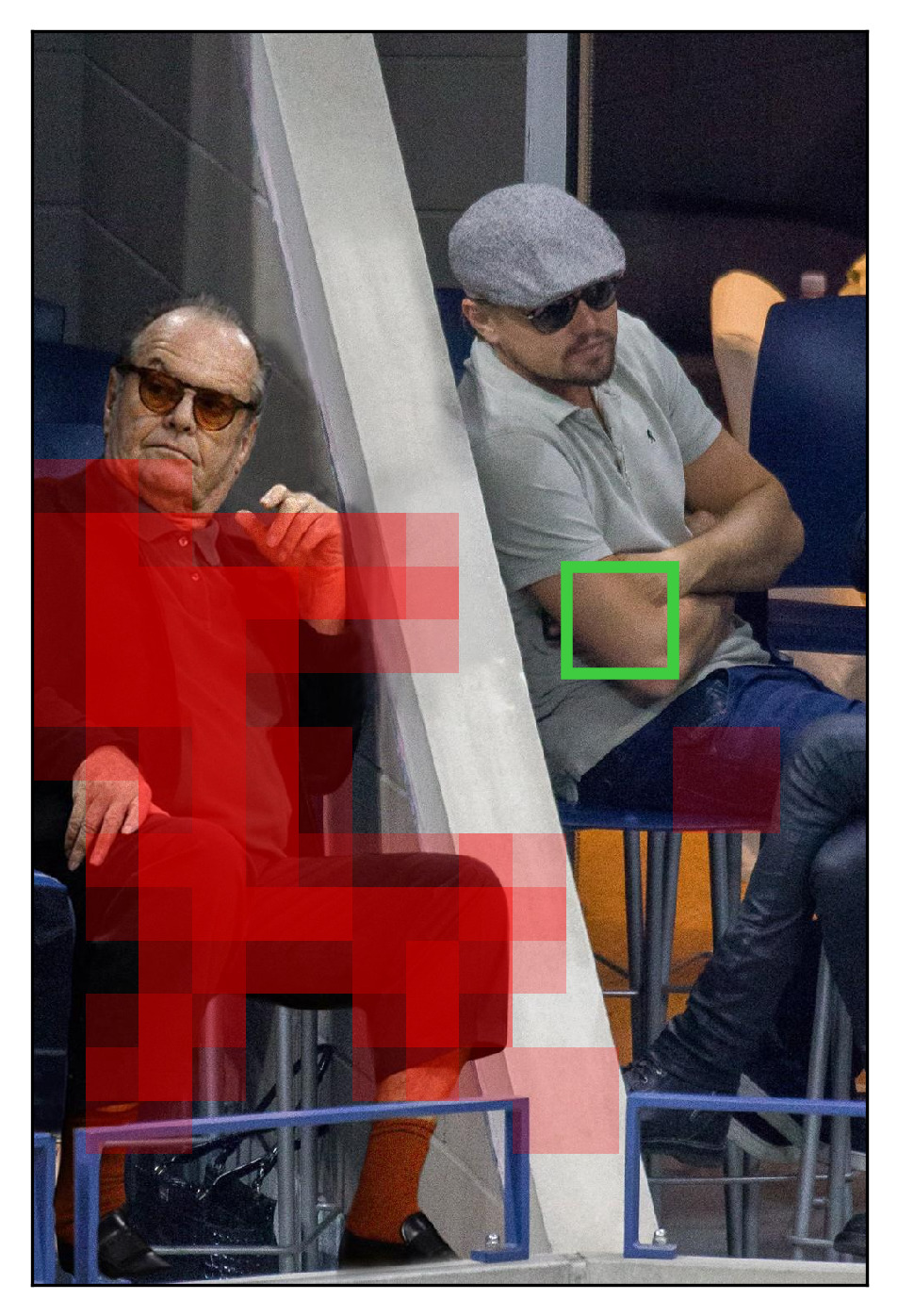}%
}
\hfill
\subfloat[\mbox{Cam. + JPG traces, 128x128}]{\enskip{\includegraphics[width=\ww]{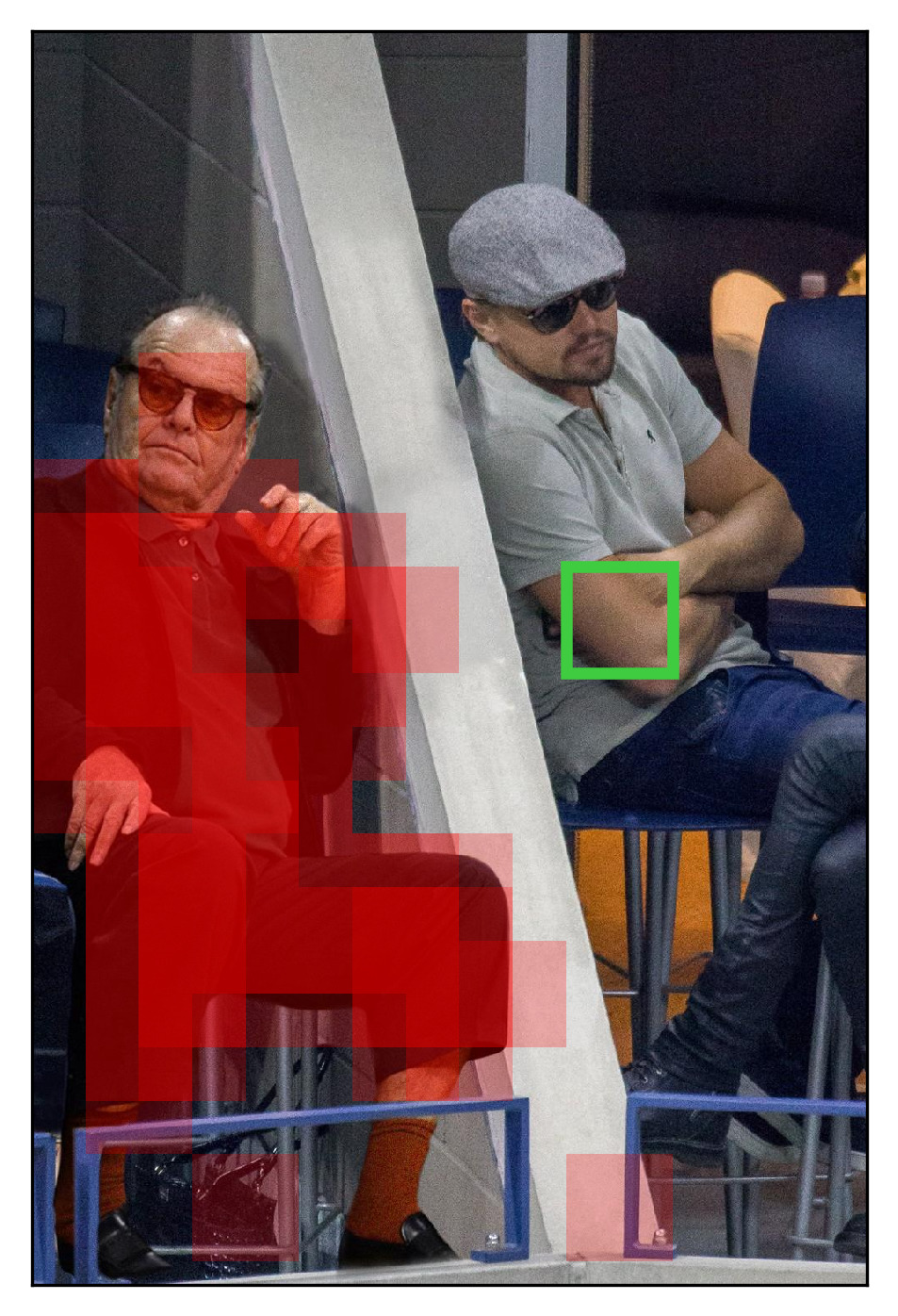}\enskip}%
}
\hfill
\subfloat[Manip. traces, 256x256]{\includegraphics[width=\ww]{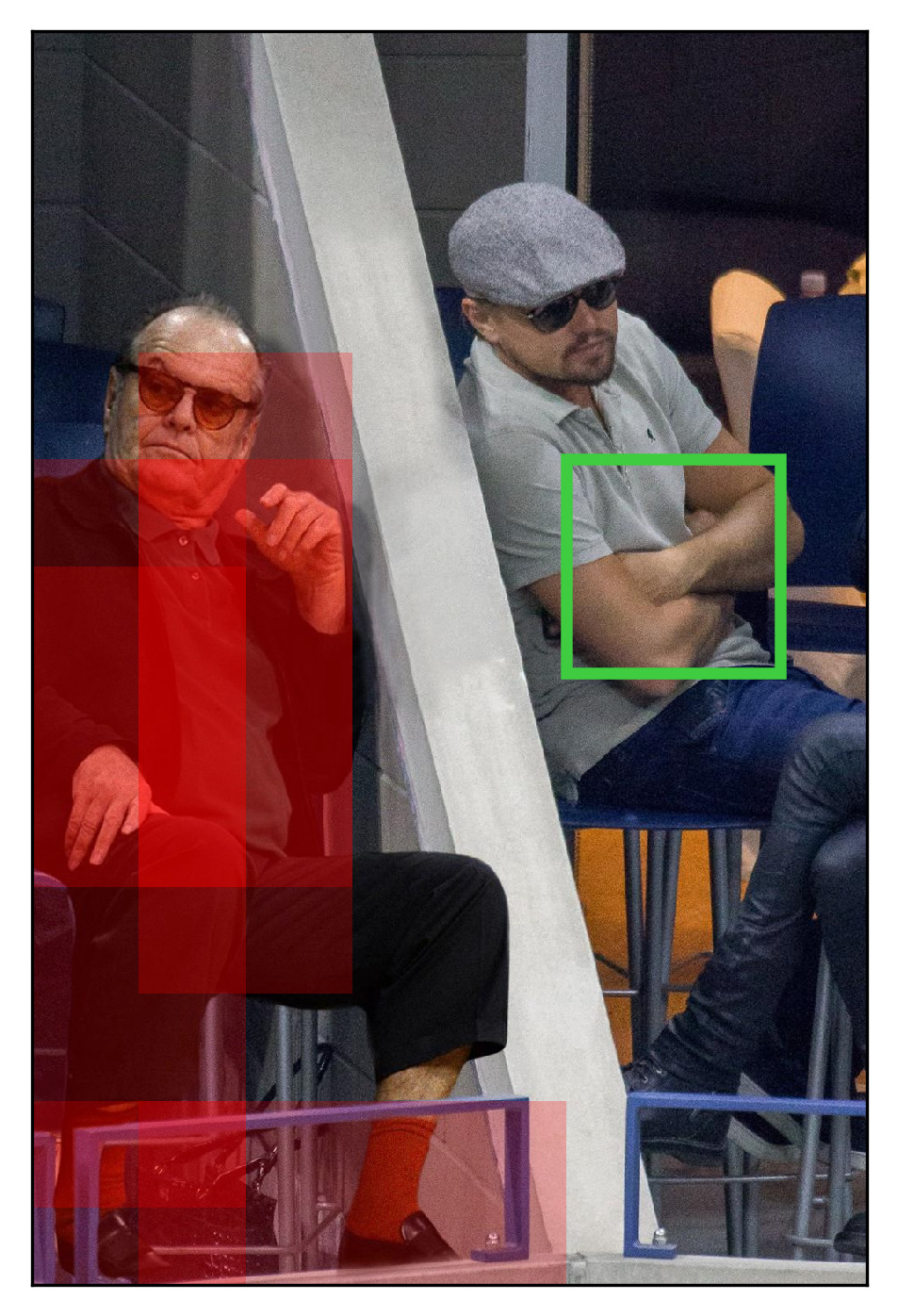}%
}
\hfill\null
\caption{Splicing detection and localization example. The green box outlines a reference patch. Patches, spanning the image with 50\% overlap, that are detected as forensically different from the reference patch are highlighted in red. Image downloaded from www.reddit.com.}
\label{fig:splice_example1}
\end{figure*}

\begin{figure*}
\centering
\null\hfill
\subfloat[Spliced Image]{\includegraphics[width=0.32\linewidth]{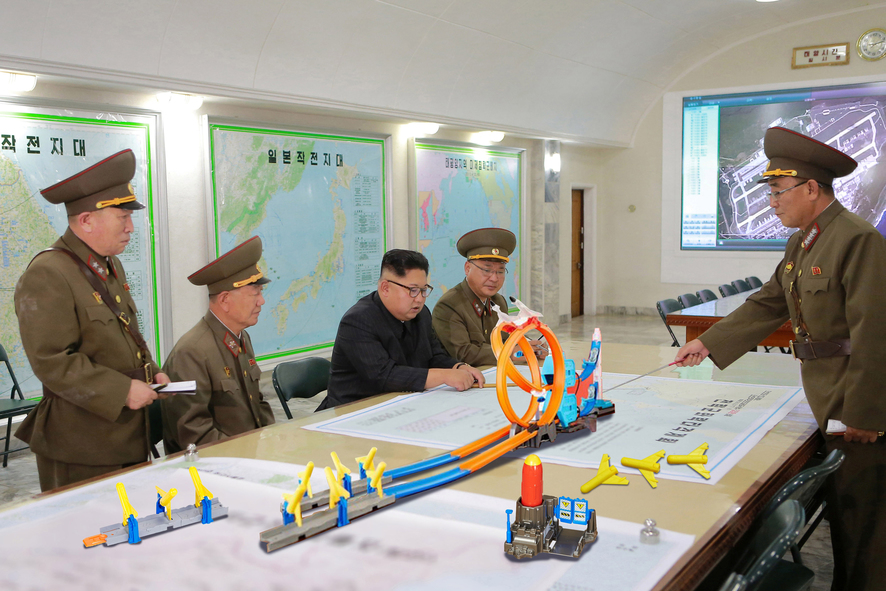}%
}
\hfill
\subfloat[Host reference, 256x256]{\includegraphics[width=0.32\linewidth]{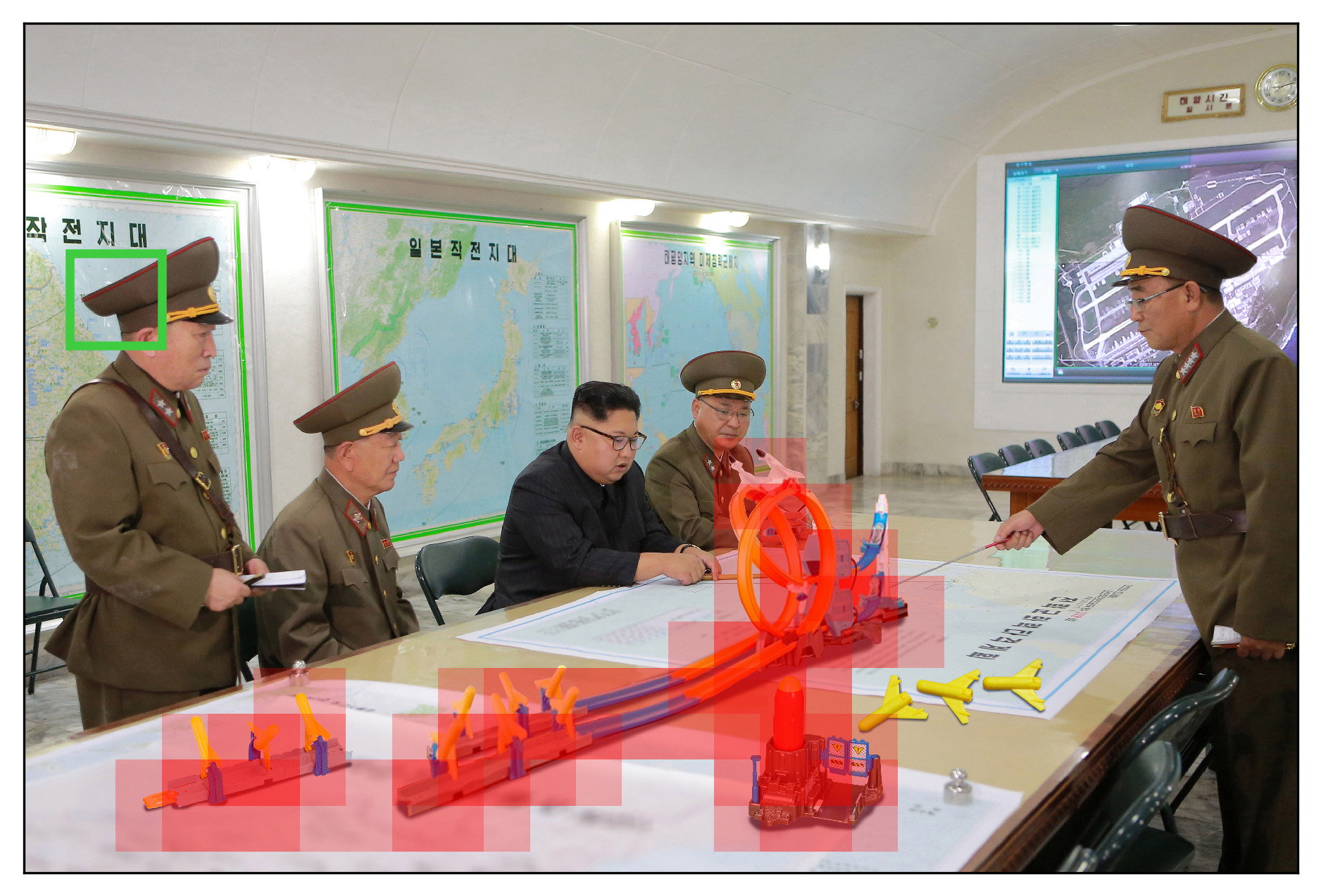}%
}
\hfill
\subfloat[Host reference, 128x128]{\includegraphics[width=0.32\linewidth]{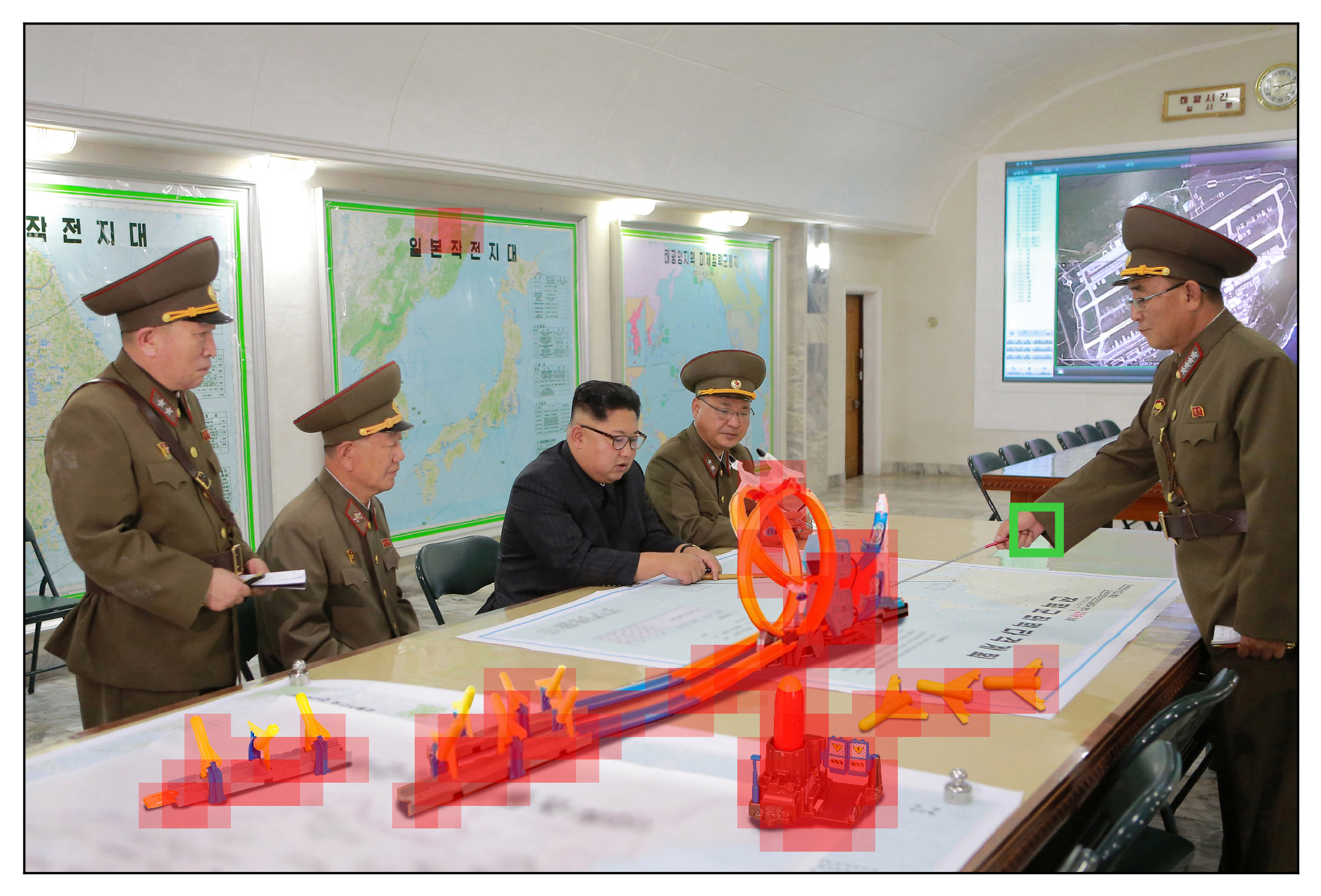}%
}
\hfill\null
\caption{Splicing detection and localization example. The green box outlines a reference patch. Patches, spanning the image with 50\% overlap, that are detected as forensically different from the reference patch are highlighted in red.}
\label{fig:splice_example2}
\end{figure*}

\begin{figure*}
\centering
\null\hfill
\subfloat[Original]{\includegraphics[width=1.5in]{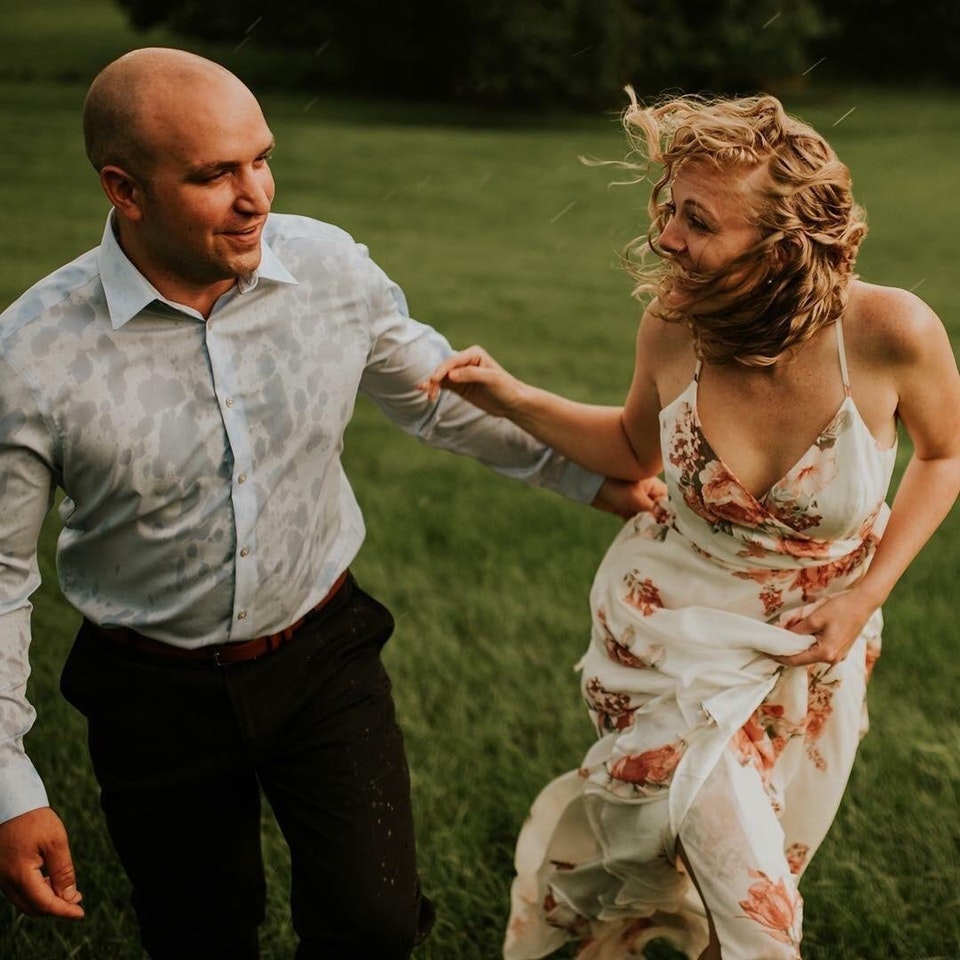}%
}
\hfill
\subfloat[Manipulated]{\includegraphics[width=1.5in]{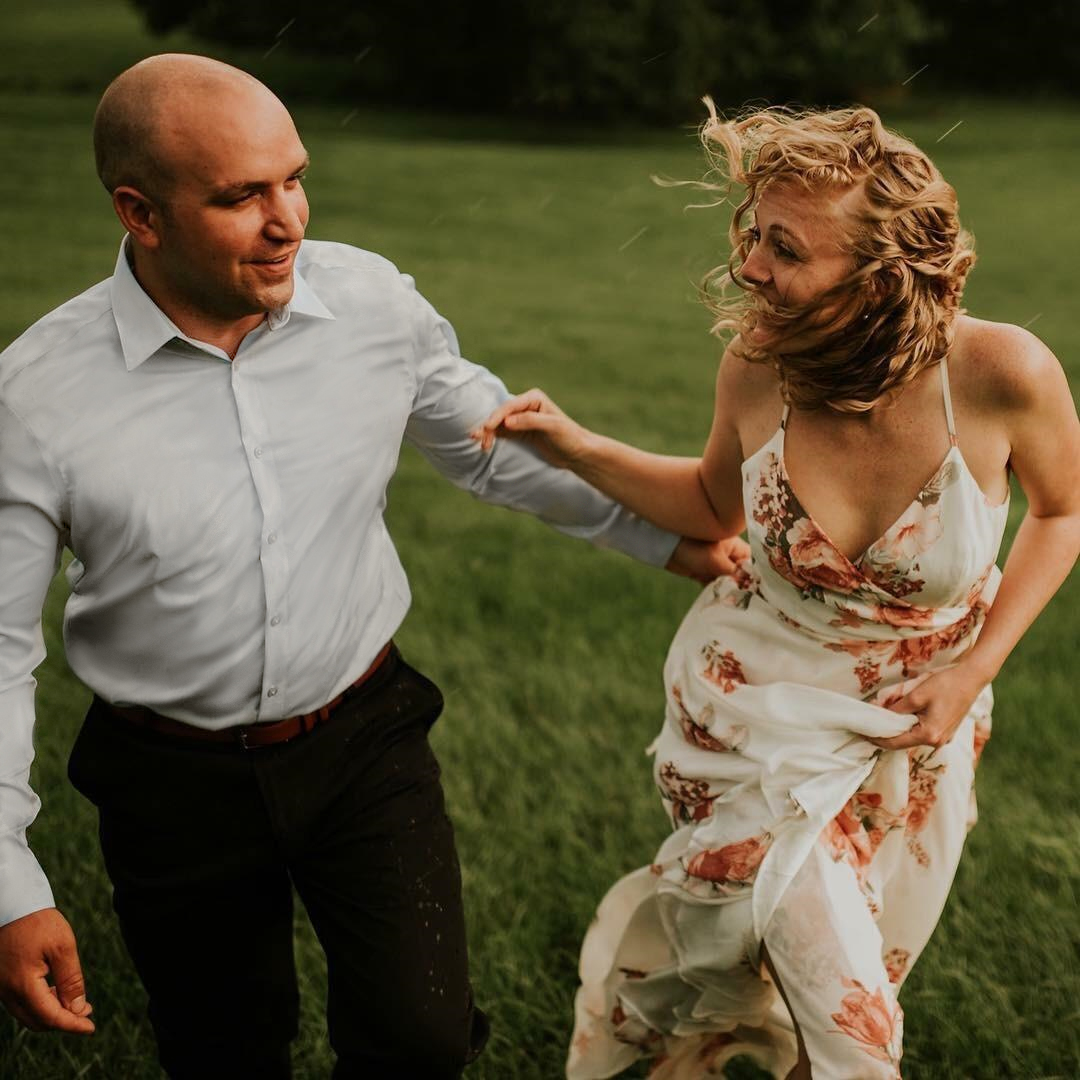}%
}
\hfill
\subfloat[Host reference]{\includegraphics[width=1.54in]{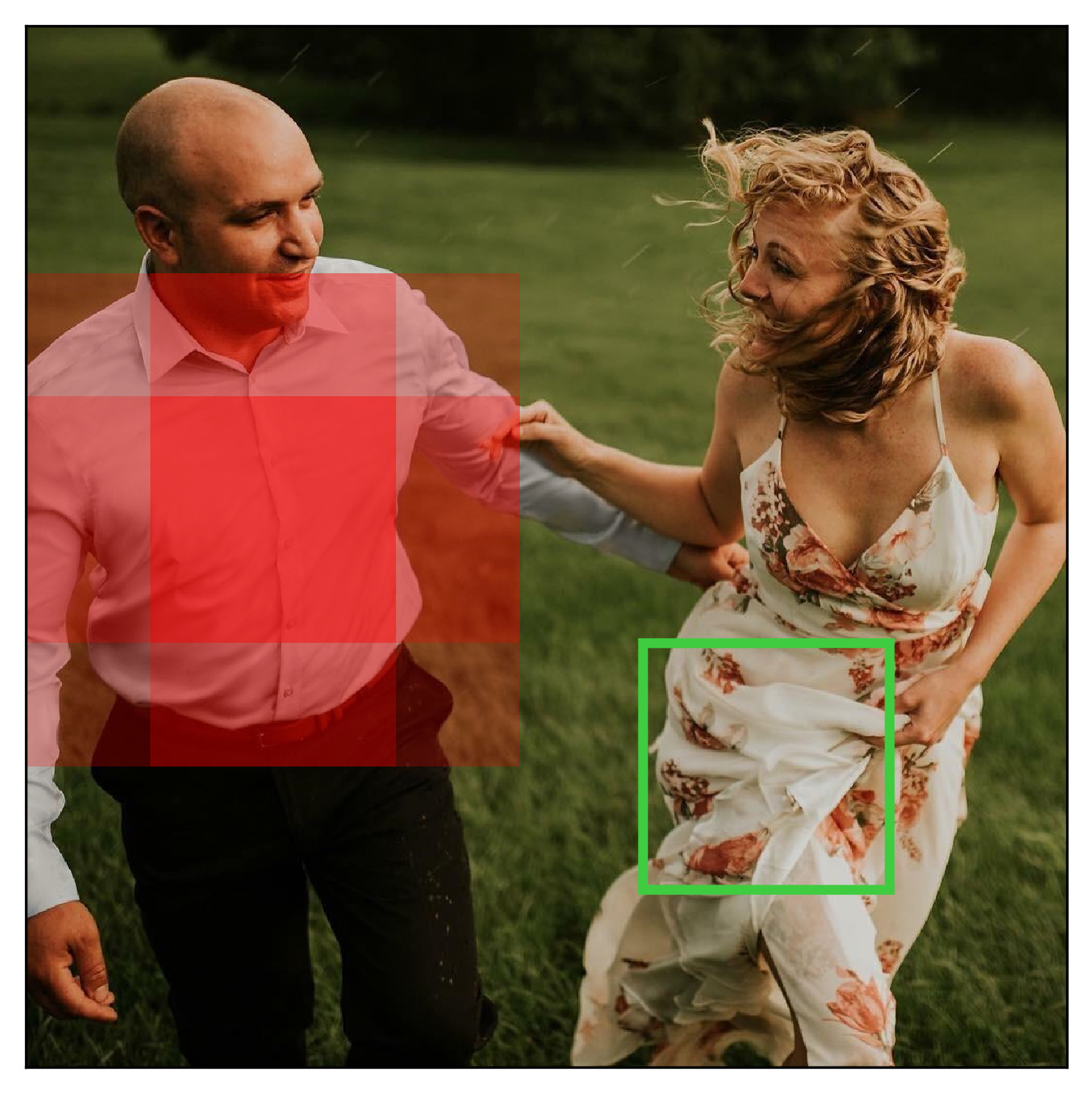}%
}
\hfill
\subfloat[Host reference, 128x128]{\includegraphics[width=1.54in]{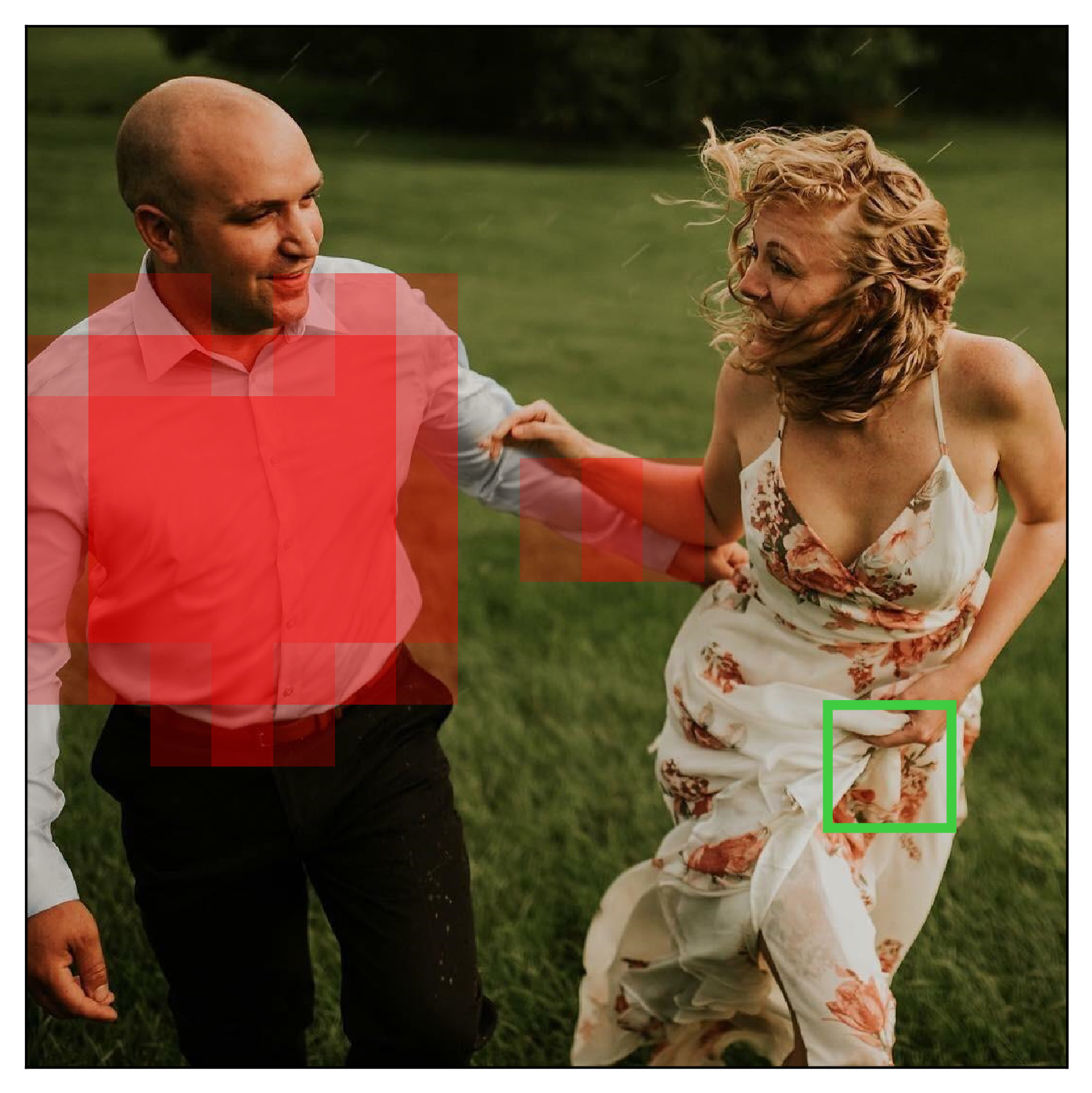}%
}
\hfill\null
\caption{Manipulation detection and localization example. The green box outlines a reference patch. Patches, spanning the image with 50\% overlap, that are detected as forensically different from the reference patch are highlighted in red.}
\label{fig:manip_example1}
\end{figure*}
%
%
%

Here we demonstrate the utility of our proposed forensic similarity system in the important forensic analysis of forged images. In forged images, an image is altered to change its perceived meaning. This can be done by inserting foreign content from another image, called splicing, or by locally manipulating a part of the image. Forging an image inherently introduces a localized inconsistency of the forensic traces in the image.\looseness=-1

In this section, we conducted two experiments that demonstrate the power of our proposed similarity system for detection of forged images and localization of the tampered region.  The proposed similarity system detects and localizes the forged region of an image by exposing that it has a different forensic trace than the rest of the image. In the first experiment, we propose a simple forgery detection criterion, evaluate its performance on three publicly available datasets, and compare against state-of-the-art forgery detection methods. In the second experiment, we show that our proposed forensic system is effective for localizing tampering in ``in-the-wild" forged images, which are visually realistic and have been downloaded from a popular social media website.\looseness=-1

To detect image forgeries, we started with three publicly available datasets. These datasets are the Columbia~\cite{hsu06crfcheck}, Carvalho~\cite{carvalho2013exposing}, and the Korus et al. ``realistic-tampering''~\cite{Korus2017TIFS} datasets, which consist of tampered and unaltered images. For each image in the dataset, subdivided it into blocks of size 128$\times$128 with 50\% overlap in each dimension. For an image with $N$ blocks, we calculated the camera-model based forensic similarity between each of $N^2$ patch pairings. We then calculated the mean similarity,
\vspace{-1mm}
\begin{equation}
m = \frac{1}{N^2}\sum^{N}_{i=1}\sum^{N}_{j=1}S(X_i,X_j),
\vspace{-1mm}
\end{equation}
where $X_i$ is the $i^{th}$ sampled block in the image, and $S(\cdot,\cdot)$ is the camera-model based similarity score from Eq.~\eqref{eq:similarity_map}. This mean similarity score is then compared to a threshold, with high values indicating no tampering occurred since all patches are expected to have high similarity with each other.

Table~\ref{tab:result_benchmark_map} shows the mean average precision for forgery detection for each dataset, with comparisons to the results achieved by the methods proposed by Huh et al.~\cite{huh2018forensics}, and Bondi et al.~\cite{bondi2017tampering}. The Huh et al. method works by generating a ``self-consistency map'' of an image, and comparing the spatial average of the consistency map to a threshold. The Bondi et al. method works by generating a authenticity membership map based on camera-model based feature representations, and comparing the spatial average of this membership map to a threshold. For~\cite{bondi2017tampering}, we use recommended settings of $\Gamma_{\text{dist}}=0.9$ and $\Gamma_{\text{conf}}=0.6$, and vary $\Gamma_{\text{det}}.$

For all three datasets, our proposed approach outperformed the Bondi et al. method and the camera-model based version of the Huh et al. method. The mean forensic similarity method achieved ``Area Under the Curve'' (AUC) scores of  0.95 on \cite{hsu06crfcheck}, 0.91 on \cite{carvalho2013exposing}, and 0.59 on~\cite{Korus2017TIFS}. There are many potential sources of these improvements, including the high performance of the proposed forensic similarity measurement, as well as differences in training data, patch size, sampling methods, and detection criteria.

Huh et al. found that training on additional information contained in the image EXIF metadata, improved performance over only camera-model information. Still, our proposed  mean camera-model based forensic similarity outperformed the Huh et al. ``EXIF'' approach in the two more challenging Carvalho and Korus datasets. This results suggests that performance may be further increased by augmenting training information to include similar EXIF-based information. This experiment demonstrates the potential power of the proposed Forensic Similarity approach for forgery detection.

\begin{table}[t]\centering
\caption{Forgery Detection Mean Average Precision}
\vspace{-1.5mm}
\setlength{\tabcolsep}{0.5em}
\begin{tabular}{l c c c} 
\toprule
Method & Columbia~\cite{hsu06crfcheck} & Carvalho~\cite{carvalho2013exposing} & Korus~\cite{Korus2017TIFS}\\
\midrule
Bondi et al.~\cite{bondi2017tampering} & 0.70 & 0.76 & 0.53 \\
Huh et al. \textit{Camera}~\cite{huh2018forensics} & 0.70 & 0.73 & 0.15 \\
Huh et al. \textit{Image}~\cite{huh2018forensics} & 0.97 & 0.75 & 0.58\\
Huh et al. \textit{EXIF}~\cite{huh2018forensics} & \textbf{0.98} & 0.87 & 0.55 \\
\midrule
\textbf{Mean Forensic Similarity} & 0.95 & \textbf{0.91} & \textbf{0.60} \\
\bottomrule
\end{tabular}
\label{tab:result_benchmark_map}
\vspace{-2.5mm}
\end{table}

In a second experiment, we demonstrate potential for forgery localization on three forged images that were downloaded from www.reddit.com, for which we also have access to the original version. First, we subdivided each forged image into image patches with 50\% overlap. Next, we selected one image patch as a reference patch and calculated the similarity score to all other image patches in the image. We used the similarity system trained in Sec.~\ref{sec:experiments1:ssec:cam_model} to determine whether two image patches were captured by the same or different camera model with secondary JPEG compression. We then highlighted the image patches with similarity scores less than a threshold, i.e. contain a different forensic trace than the reference patch.\looseness=-1

Results from this procedure on the first forged image are shown in Fig.~\ref{fig:splice_example1}. The original image is shown in Fig.~\ref{fig:splice_example1}a. The spliced version is shown Fig.~\ref{fig:splice_example1}b, where an actor was spliced into the image. 
When we selected a reference patch from the host (original) part of the image, the image patches in the spliced regions were highlighted as forensically different as shown in Fig.~\ref{fig:splice_example1}c.
We note that our forensic similarity based approach is agnostic to which regions are forged and which are authentic, just that they have different forensic signatures. This is seen in Fig.~\ref{fig:splice_example1}d when we selected a spliced image patch as the reference patch. Fig.~\ref{fig:splice_example1}e shows then when we performed this analysis on the original image, our forensic similarity system does not find any forensically different patches from the reference patch.\looseness=-1

The second row of Fig.~\ref{fig:splice_example1} shows forensic similarity analysis using networks trained under different scenarios. Results using the network trained to determine whether two image patches have the same or different source camera model without JPEG post-compression are shown in Fig.~\ref{fig:splice_example1}f for patch size 256$\times$256, in Fig.~\ref{fig:splice_example1}g for patch size 128$\times$128, and with JPEG post-compression in Fig.~\ref{fig:splice_example1}h for patch size 128$\times$128. The result using the network trained to determine whether two image patches have been manipulated by the same or different manipulation type is shown in Fig.~\ref{fig:splice_example1}i.

Results from splicing detection and localization procedure on a second forged image are shown in Fig.~\ref{fig:splice_example2}, where a set of toys were spliced into an image of a meeting of government officials. When we selected reference patches from the host image, the spliced areas were correctly identified as containing a different forensic traces, exposing the image as forged. This is seen in Fig.~\ref{fig:splice_example2}b with 256$\times$256 patches, and in Fig.~\ref{fig:splice_example2}c with 128$\times$128 patches. The 128$\times$128 case showed better localization of the spliced region and additionally identified the yellow airplanes as different than the host image, which were not identified by the similarity system using larger patch sizes.\looseness=-1

In a final example, shown in Fig.~\ref{fig:manip_example1}, the raindrop stains on a mans shirt were edited out the image by a forger using a brush tool. When we selected a reference patch from the unedited part of the image, the manipulated regions were identified as forensically different, exposing the tampered region of the image. This is seen in Fig.~\ref{fig:manip_example1}c with 256$\times$256 patches, and in Fig.~\ref{fig:manip_example1}d with 128$\times$128 patches. In the 128$\times$128 case, the smaller patch size was able to correctly expose that the man's shirt sleeve was also edited. 

The result in Fig.~\ref{fig:manip_example1} is particularly interesting since the image has been modified by a brushing operation, yet it is being detected by a network trained for comparisons of source camera model. Research in~\cite{mayer2018unified} showed that forensic features learned for camera model identification can be used to detect different manipulations with high accuracy. Furthermore, results in~\cite{cozzolino2018noiseprint} showed that their forgery localization technique trained with camera model features was able to localize in-painting manipulations. These results suggest that the forensic traces induced by the source camera model and forensic traces induced by manipulations are related, which is corroborated by our result in Fig.~\ref{fig:manip_example1}.

The experiments in this section show that the proposed forensic similarity based approach is a powerful technique for detection and localization of image forgeries. The experiments showed that correctly identifying differences in forensic traces was sufficient to expose the forgery and forgery locations, even though the technique did not identify the particular forensic traces in the image.

\vspace{-2mm}
\subsection{Database Consistency Verification}
\begin{table}[t]\centering
\caption{Rates of correctly identifying a database as ``consistent" (i.e. all the same camera model) or ``inconsistent,'' M=10, N=20}
\vspace{-2mm}
\label{tab:database_consistency}
\begin{tabular}{l l r r r} 
\toprule
Recompression & threshold & Type 0 & Type 1 & Type 2\\
\midrule
None & 0.1 & 97.4\% & 84.8\% & 100\% \\
 & 0.5 & 92.4\% & 91.9\% & 100\% \\
 & 0.9 & 70.7\% & 96.7\% & 100\% \\
\midrule
QF=75 & 0.1 & 99.2\% & 18.5\% & 92.2\% \\
& 0.5 & 75.9\% & 75.5\% & 100\% \\
& 0.9 & 00.2\% & 100\% & 100\% \\
\bottomrule
\end{tabular}
\vspace{-3mm}
\end{table}

In this section, we demonstrate that the forensic similarity system detects whether a database of images has either been captured by all the same camera model, or by different camera models. This is an important task for social media websites, where some accounts illicitly steal copyrighted content from many different sources. We refer to these accounts as ``content aggregators", who upload images captured by many different camera models. This type of account contrasts with ``content generator" accounts, who upload images captured by one camera model. In this experiment, we show how forensic similarity is used to differentiate between these types of accounts.\looseness=-1

To do this, we generated three types of databases of images. Each database contained $M$ images and were assigned to one of three ``Types." Type 0 databases contained $M$ images taken by the same camera model, i.e. a content generator database. Type 1 databases contained $M-1$ images taken by one camera model, and 1 image taken by a different camera model. Finally, Type 2 databases contain $M$ images, each taken by different camera models. We consider the Type 1 case the hardest to differentiate from a Type 0 database, whereas the Type 2 case is the easiest to detect. We created 1000 of each database type from images captured by camera models in set $\mathcal{C}$, i.e. unknown camera models not used in training, with the camera models randomly chosen for each database. 
We then created duplicate versions of these databases, where each image is re-compressed with a JPEG quality factor of 75. This was done to mimic conditions similar to images on social media websites.

To classify a database as consistent or inconsistent, we examined each ${M \choose 2}$  unique image pairings of the database. For each image pair, we randomly selected $N$ 256$\times$256 image patches from each image and calculated the $N^2$ similarity scores across the two images. Similarity was calculated using the similarity network trained in Sec.~\ref{sec:experiments1:ssec:cam_model}. Then, we calculated the median value of scores for each whole-image pair. For image pairs captured by the same camera model this value is high, and for two images captured by different camera models this value is low. We then compare the $(M-2)$th lowest value calculated from the entire database to a threshold. If this $(M-2)$th lowest value is above the threshold, we consider the database to be consistent, i.e. from a content generator. If this value is below the threshold, then we consider the database to be inconsistent, i.e. from a content aggregator.

Table~\ref{tab:database_consistency} shows the rates at which we correctly classify Type 0 databases as ``consistent" (i.e. all from the same camera model) and Type 1 and Type 2 databases as ``inconsistent'', with $M=10$ images per database, and $N=20$ patches chosen from each image. Under the no re-compression case and at a threshold of 0.5, we correctly classified 92.4\% of Type 0 databases as consistent, and correctly classified 91.9\% of Type 1 databases as inconsistent. This incorrect classification rate of Type 1 databases is decreased by increasing the threshold. Even at a very low threshold of 0.1, our system correctly identified all Type 2 databases as inconsistent. 

Under the case where all images have undergone JPEG re-compression, and at a threshold of 0.5, we correctly classified 75.9\% of Type 0 databases as consistent, correctly classified 75.5\% of Type 1 databases as inconsistent, and correctly classified 100\% of Type 2 databases as inconsistent.

The results of this experiment show that our proposed forensic similarity system is effective for verifying the consistency of an image database, an important type of forensic investigation. Importantly, the forensic similarity system was effective for evaluating consistency of databases captured by camera models not used to train the system.

\section{Conclusion}
In this paper we proposed a new digital image forensics technique, called forensic similarity, which determines whether two image patches contain the same or different forensic traces. The main benefit of this approach is that prior knowledge, e.g. training samples, of a forensic trace are not required to make a forensic similarity decision on it. To do this, we proposed a two part deep-learning system composed of a CNN-based feature extractor and a three-layer neural network, called the similarity network, which maps pairs of image patches onto a score indicating whether they contain the same or different forensic traces. We experimentally evaluated the performance of our approach on three types of common forensic scenarios, which showed that our proposed system was accurate in a variety of settings. Importantly, the experiments showed this system is accurate even on ``unknown" forensic traces that were not used to train the system and that our proposed system significantly improved upon prior art in~\cite{mayer2018similarity}, reducing error rates by over 50\%. Furthermore, we demonstrated the utility of the forensic similarity approach in two practical applications of forgery detection and localization, and image database consistency verification.

\ifCLASSOPTIONcaptionsoff
  \newpage
\fi

\bibliographystyle{IEEEtran}
\bibliography{bib/mmf,bib/mmf_deeplearning,bib/cv_deepfeatures,bib/cv_deeplearning}

\end{document}